\documentclass[11pt]{article}
\usepackage{amsmath}
\usepackage[preprint]{acl}

\usepackage{times}
\usepackage{latexsym}
\usepackage{booktabs}
\usepackage[T1]{fontenc}

\usepackage[utf8]
{inputenc}
\usepackage{enumitem}

\usepackage{microtype}

\usepackage{inconsolata}
\usepackage{dsfont}
\usepackage{amsfonts}
\newcommand{\indicator}{\mathbb{1}}
\DeclareMathOperator*{\argmax}{arg\,max}
\DeclareMathOperator{\Top}{Top}
\usepackage{amsmath}
\usepackage{amssymb}

\newcommand{\Nup}{\mathcal{N}^{+}}
\newcommand{\Ndown}{\mathcal{N}^{-}}
\newcommand{\Rup}{\mathcal{R}^{+}}
\newcommand{\Rdown}{\mathcal{R}^{-}}
\newcommand{\Lup}{\mathcal{L}^{+}}
\usepackage{graphicx}

\title{Reading Between the Layers: Where Representation Quality Lives, and What Geometry Reveals}

\author{\textbf{Arnesh Batra\textsuperscript{1,*}, Aniket Khandelwal\textsuperscript{1,*,$\ddagger$}, Arush Gumber\textsuperscript{1,*}, Krish Thukral\textsuperscript{2}} \\
  \textbf{Kushagra Agrawal\textsuperscript{1,$\dagger$}, Dev Sharma\textsuperscript{1,$\dagger$}} \\
  \textsuperscript{1}Indraprastha Institute of Information Technology, Delhi \\
  \textsuperscript{2}Manipal University Jaipur}

\begin{document}
\maketitle
\begingroup
\renewcommand{\thefootnote}{}
\footnotetext{\textsuperscript{*}Equal contribution.\quad\textsuperscript{$\dagger$}Equal-third authors.\\
\textsuperscript{$\ddagger$}Corresponding author: \href{mailto:aniket23087@iiitd.ac.in}{\texttt{aniket23087@iiitd.ac.in}}.\\
Code is available on GitHub at \url{https://github.com/arnesh2212/Reading_Between_Layers}.}
\endgroup
\begin{abstract}
Representation quality in pretrained text foundation models is not concentrated at the final layer, yet practitioners routinely probe it there because layer selection without labels has remained an open problem. We present a large-scale empirical study of where layer quality lives across 25 models spanning four architectural families and four pretraining strategies, evaluated on 13 datasets across classification, pair classification, semantic textual similarity, and multiple-choice question answering. Peak probing layers depend systematically on pretraining strategy and architecture: masked encoders peak in late-middle layers, sequence-to-sequence models shift their peak with task type, contrastive sentence encoders concentrate signal at the final layer under CLS and last-token readouts but peak early under mean pooling, and decoder-only models exhibit distinct phase structure tied to depth. We distill these findings into the Stable Geometry Index (SGI), a closed-form, fully label-free selector with no fitted coefficients combining power-law fit quality, effective-rank capacity, spectral stability, and a weak depth prior. Across 383 evaluation groups SGI improves over the last-layer baseline and lifts Hit@3 from 0.39 to 0.52, recovering 40\% of the gap to the oracle without using any labels, and exceeds a supervised ELMo-style layer mix on the matched subset despite using strictly less information. Our ablations also reverse a common assumption in the spectral-decay literature: power-law fit quality is a stronger predictor of layer quality than the exponent's proximity to one.
\end{abstract}

\section{Introduction}
\label{sec:introduction}

\vspace{-0.75em}

Pretrained text models are usually probed at the final layer, but representation quality is not concentrated there.
By representation quality we mean downstream utility or transferability, operationalised by performance on the evaluated downstream tasks under our fixed probing protocol. Geometry provides a label-free selection signal; downstream performance defines quality.
Prior work shows that linguistic, syntactic, and transfer-relevant information varies across depth, with intermediate layers often carrying stronger signal than the final layer \citep{tenney2019bert, liu2019linguistic, voita2019bottom, fan2024not, lad2024remarkable}. The problem is that the best layer is not known in advance. In practice, choosing it usually requires a labeled sweep over all layers, which is expensive and defeats the purpose of low-label probing.
This paper asks two questions. \textbf{Where does representation quality live across model families, pretraining objectives, and task types?} \textbf{Can we locate the best layer from unlabeled geometry alone?}

We first study 25 pretrained models across four broad architectural families: encoders, encoder--decoders, decoder-only models, and state-space models. Sentence-transformer models are treated as encoder-family models because they share an encoder-style representation interface. The study covers 11 pretraining objectives or objective variants and 13 datasets across four task categories. Across all peak-layer instances, best layers are strongly structured by task, readout, and pretraining objective. Mean pooling peaks early in 61.1\% of runs, while CLS readouts peak late in 66.8\%. Sequence-to-sequence models peak early on classification but later on multiple-choice QA, and contrastive sentence encoders concentrate signal near the final layer under CLS and last-token readouts while peaking early under mean pooling.

We then introduce the \textbf{Stable Geometry Index (SGI)}, a closed-form, fully label-free layer selector computed from a fraction of unlabeled embeddings. SGI combines four signals: power-law fit quality, effective-rank capacity, spectral stability, and a weak late-middle depth prior. The rule is additive and uses no learned coefficients. Across 383 encoder evaluation groups, SGI improves over the last-layer baseline by 1.20 percentage points and raises Hit@3 from 0.389 to 0.517, recovering 40\% of the oracle gap without using labels. SGI improves over popular label-free geometry baselines, including effective rank, RankMe \citep{garrido2023rankme}, IsoScore \citep{rudman2022isoscore}, participation ratio, intrinsic dimension, and entropy. The best single-signal Hit@3 among these baselines is 0.428, while SGI reaches 0.517. On the matched CLS subset, SGI also exceeds a supervised ELMo-style learned layer mix \citep{peters2018deepcontextualizedwordrepresentations} by 1.41 metric points, despite using strictly less information.

Our contributions:
\begin{itemize}[leftmargin=*, noitemsep, topsep=0pt]

\item \textbf{A large-scale layerwise empirical study.} We characterize representation quality across 25 models, 11 pretraining objectives or variants, 13 datasets, four task categories, and four pooling strategies.
\item \textbf{A cross-family analysis of layer structure.} We show that peak layers are not arbitrary: they vary systematically with architecture, pretraining objective, task type, and readout strategy.
\item \textbf{The Stable Geometry Index (SGI).} We propose a closed-form, label-free geometric selector that improves over the final layer, beats single-signal geometric baselines, and exceeds a supervised learned layer mix on the matched subset.
\item \textbf{A label-free pruning prior.} For $22\%$ of encoder evaluations, SGI identifies a first-quintile cutoff layer where the upper $80\%$ of layers can be removed, yielding a $5\times$ inference reduction while improving the probe metric.

\end{itemize}

\vspace{-0.5em}

\section{Related Works}
\label{sec:related}
\vspace{-0.5em}

\subsection{Intermediate Layer Representations}
\label{subsec:intermediate}
\vspace{-0.5em}

Prior work shows that transformer representations are distributed
non-uniformly across layers, and final-layer embeddings are often not
optimal. In encoder models such as BERT, linguistic features are
frequently best captured in middle layers rather than final layers
~\citep{liu2019linguistic, tenney2019bert, voita2019bottom}. Similar
results in large language models show that mid-depth layers often encode
more robust and transferable features
~\citep{gurnee2023language, lad2024remarkable,
fan2024not}. Related work on \emph{attention sinks} shows that attention
can collapse onto a small set of tokens
~\citep{attention-sinks, identifiability, gu2024attention}, while
intermediate decoder layers exhibit weaker sink behavior than shallow or
deep layers ~\citep{first-token-attending}.
\vspace{-0.5em}

\subsection{Geometry and Layer-wise Representation Quality}
\label{subsec:geometry_metrics}
\vspace{-0.5em}

A broad line of work studies embedding geometry as a label-free proxy
for representation quality. Isotropy and anisotropy have been widely
analyzed in transformer embeddings
~\citep{mu2018allbutthetop,cai2021isotropy,ethayarajh2019contextual},
with locally isotropic token-type clusters observed in contextual models
such as BERT and ERNIE ~\citep{rajaee2021isotropy}. Several diagnostics
quantify these properties, including anisotropy and isotropy-based
measures ~\citep{bihani2021lowanisotropy,rudman2022isoscore},
Participation Ratio, stable rank, and spectral
methods such as power-law exponents for OOD generalization
~\citep{martin2021implicitselfreg} and RankMe, which estimates
effective rank via singular-value entropy
~\citep{garrido2023rankme}. Other geometric diagnostics, including
LiDAR’s covariance-based score ~\citep{thilak2024lidar}, Jacobian
effective rank ~\citep{chung2026global}, manifold distortion measures, and
intrinsic dimensionality analyses of fine-tuning dynamics
~\citep{aghajanyan2021intrinsic}, further connect geometry to
downstream utility. Building on these signals, several methods evaluate intermediate
representations without downstream labels. LevyScore
~\citep{maes2024levyscore} measures confidence via deviations from
pretraining embedding density, while Layer by Layer
~\citep{skean2025layerbylayer} uses entropy, curvature, and
augmentation-invariance to show that mid-depth layers often outperform
final layers. Perception Encoder
~\citep{bolya2025perception} similarly identifies strongest visual
embeddings at intermediate depths, while
~\citep{balestriero2025lejepa} theoretically relate isotropic Gaussian
structure to lower downstream risk. In language models,
~\citep{cheng2025emergence,razzhigaev2024shape} show that abstraction,
intrinsic dimension, and anisotropy evolve systematically with depth
and correlate with probing performance. LOES
~\citep{batra2026uncoveringlatentpotentialdeep} further formulates
layer selection as spectral residual optimization under orthogonality
and isotropy constraints. Together, these works motivate universal
layer-selection methods across models and tasks.
\vspace{-0.75em}

\section{Methodology}
\label{sec:methodology}
\vspace{-0.75em}

Transformer encoders expose a sequence of representation spaces rather than a single representation.  Prior layer-wise analyses show that linguistic and transfer information is distributed across depth rather than concentrated only at the final layer \citep{tenney2019bert,liu2019linguistic,jawahar-etal-2019-bert}. The methodological question is therefore: can we choose a useful layer from unlabeled geometry alone, before seeing any downstream probe score?  SGI answers
this by treating each layer as a point cloud and scoring the shape of its spectrum and local neighborhood geometry.
The geometric intuition is simple: a useful transfer layer should be broad enough to encode many directions, stable enough not to collapse onto a few dominant axes, locally spread enough to preserve neighborhood structure, and deep enough to move beyond shallow lexical features.  SGI scores these properties with the same fixed expression for every task, readout, model, and split.

\vspace{-0.5em}

\subsection{Geometry of a Layer}
\label{subsec:layer-geometry}
\vspace{-0.5em}

At layer $\ell$, let $H_\ell \in \mathbb{R}^{n_\ell \times d}$ be the matrix of representations with singular values $\sigma_{\ell,1} \geq \cdots \geq \sigma_{\ell,q} > 0$. We characterise the layer geometry through four spectral and manifold quantities, each grounded in a distinct theoretical principle: spectral entropy as a measure of capacity, deviation from a single power law as a measure of structured residual information, spectral conditioning as a measure of stability, and local neighbourhood scaling as a measure of intrinsic dimensionality. Together these capture the global, residual, conditioning, and local-geometric properties of the representation manifold.

\vspace{-0.5em}
\paragraph{Spectral capacity.}
Following \citet{roy2007effective} and \citet{garrido2023rankme}, we measure how broadly information is spread across singular directions via the exponentiated Shannon entropy of the normalised spectrum,
\vspace{-1em}

\[
  \pi_{\ell,i} = \frac{\sigma_{\ell,i}}{\sum_j \sigma_{\ell,j}}, \;
  e_\ell = \exp\!\left(-\sum_i \pi_{\ell,i} \log \pi_{\ell,i}\right).
\]
\vspace{-1em}

This quantity is invariant to scalar rescaling of $H_\ell$ and equals the number of directions a maximum-entropy distribution with the same support would occupy~\citep{garrido2023rankme}, giving it a clean information-theoretic interpretation as the effective rank of the representation.
If $k$ singular directions carry equal mass and the rest are negligible, then $e_\ell=k$. It therefore measures spectral breadth rather than hidden-state width. In a linear probe, producing a fixed logit change along a small-singular-value direction requires a larger coefficient, which can amplify finite-sample noise. Broad spectral support can make more directions numerically usable, but does not establish that they separate the downstream labels.
\vspace{-0.5em}

\paragraph{Spectral residual structure.}
Heavy-tailed singular spectra are predicted by random-matrix theory and have been linked to implicit self-regularisation in trained networks~\citep{martin2019traditionalheavytailedselfregularization,martin2021implicitselfreg}. We regress $\log \sigma_{\ell,i}$ on $\log i$ and use the coefficient of determination $x_\ell$ as a measure of how well a single power law explains the spectrum. Low $x_\ell$ indicates structured deviation from the heavy-tailed regime, which (as our construction ablations confirm in Section~\ref{subsec:ablations}) carries more layer-quality signal than the exponent value itself.
\vspace{-0.5em}
\paragraph{Stability.}
The condition number $\kappa_\ell = \sigma_{\ell,1} / \sigma_{\ell,q}$ measures spectral ill-conditioning: a nearly degenerate tail can make fitting a linear probe sensitive to perturbations, although actual sensitivity also depends on the labels, loss, and regularisation. Large condition numbers are consistent with the anisotropy and rogue-dimension phenomena observed in contextual encoders~\citep{ethayarajh2019contextual,timkey-van-schijndel-2021-bark}. Conditioning thus complements effective rank by detecting weak directions that a broad global spectrum can obscure.
\vspace{-0.5em}

\paragraph{Local intrinsic spread.}
The TwoNN estimator~\citep{Facco2017} recovers the intrinsic dimensionality of the data manifold from the ratio of first- and second-nearest-neighbour distances under a locally Poisson assumption, yielding a consistent estimator that is invariant to the embedding dimension $d$. Local intrinsic dimensionality complements the global spectrum: $e_\ell$ describes how variance is distributed across orthogonal directions, while $\delta_\ell$ describes how points populate the manifold in their immediate neighbourhood.
\vspace{-0.5em}

\subsection{Group-wise normalisation}
\label{subsec:normalization}

To keep the selector invariant to model- and dataset-specific scales, every trace is normalised across the layers of a single evaluation group. We use three normalisers, each chosen for a specific statistical property of the underlying trace. The min-max normaliser
\vspace{-1em}

\begin{align*}
  \Nup(u)_\ell &= \frac{u_\ell - \min_j u_j}{\max_j u_j - \min_j u_j},\\
  \Ndown(u)_\ell &= 1 - \Nup(u)_\ell,
\end{align*}
\vspace{-1em}

is appropriate for bounded, well-behaved traces (constant traces map
to $0.5$). The rank normaliser
\begin{align*}
  \Rup(u)_\ell &= \frac{\operatorname{rank}(u_\ell) - 1/2}{L},\\
  \Rdown(u)_\ell &= 1 - \Rup(u)_\ell,
\end{align*}
uses average ranks for ties and is robust to outliers in heavy-tailed traces. The log-min-max normaliser
\[
  \Lup(u)_\ell = \Nup\!\left(\log(1 + u_\ell - \min_j u_j)\right),
\]
applies a variance-stabilising transform before scaling and is the appropriate choice for traces with multiplicative dynamic range such as effective rank and intrinsic dimension. 
\vspace{-0.5em}

\subsection{Stable Geometry Index}
\label{subsec:sgi}

We combine the complementary geometric properties above in a fixed additive score, the Stable Geometry Index:
\vspace{-1em}

\[
  G_\ell = R_\ell + C_\ell + S_\ell + P_\ell.
\]
\vspace{-1em}

The additive form is a design choice supported by the ablations in Section~\ref{subsec:ablations}, Table~\ref{tab:ablation}: removing any coordinate or replacing the stability--depth sum with a product reduces performance.

The \emph{residual} term $R_\ell = \Ndown(x)_\ell^2$ rewards spectra whose deviation from a power law is large in absolute terms; the square sharpens the contribution of layers near the maximum residual. The \emph{capacity} term
\[
  Z_\ell = \Lup(e)_\ell, \qquad C_\ell = Z_\ell (1 + Z_\ell),
\]

rewards broad usable rank, with the quadratic factor $(1 + Z_\ell)$ giving a steeper gradient near saturation. The \emph{stability} term aggregates value-, rank-, and manifold-side conditioning,
\[
  S_\ell = \Ndown(\kappa)_\ell + \Rdown(\kappa)_\ell + \Lup(\delta)_\ell,
\]
preferring layers that are well-conditioned in both spectral value and ordering while retaining local intrinsic spread. The \emph{depth prior}
\[
  P_\ell = \Nup\!\left(d_\ell^4 (1 - d_\ell)\right), \qquad d_\ell = \ell / L,
\]
encodes an empirically supported prior over depth: probing accuracy in pretrained transformers peaks in late-middle layers across linguistic, syntactic, and transfer probes~\citep{jawahar-etal-2019-bert,tenney2019bert,liu2019linguistic}. The functional form $d^4(1-d)$ places its mode at $d = 4/5$, a fixed late-middle location consistent with the depth at which prior probing studies report peak performance. The prior is bounded and weak: it perturbs the ranking of geometrically near-tied layers without overriding strong spectral evidence at other depths.

The selected layer is the latest maximiser, $\hat\ell = \max\argmax_\ell G_\ell$, where the tie rule is deterministic and favours deeper layers only when the geometry is otherwise indistinguishable.
\vspace{-0.5em}

\section{Experimental Setup}
\vspace{-0.5em}

\label{sec:experiments}
\subsection{Models}
\vspace{-0.5em}

\label{sec:models}
Our main evaluation covers 25 pretrained models spanning four architectural families and a range of pretraining objectives, positional encodings, and parameter scales (66M--581M). A separate scaling probe adds four checkpoints up to 4B parameters (Section~\ref{subsec:scaling}).
\vspace{-0.5em}

\paragraph{Encoders (MLM and variants).}
Our primary cohort is 12 bidirectional encoders trained with masked language modelling or related discriminative objectives: BERT~\citep{devlin2019bert} (base uncased, large uncased, large whole-word-masked), DistilBERT~\citep{sanh2019distilbert}, RoBERTa-base~\citep{liu2019roberta}, DeBERTa-v3-base~\citep{he2021debertav3}, ModernBERT-base~\citep{warner2025smarter}, SpanBERT-base~\citep{joshi2020spanbert}, and ERNIE-2.0-base~\citep{sun2020ernie}. We also add multilingual BERT~\citep{devlin2019bert}, XLM-RoBERTa-base~\citep{conneau2020unsupervised}, and XLM-MLM-100~\citep{conneau2019cross}.
\vspace{-0.5em}

\paragraph{Encoders (contrastive sentence objective).}
Four encoders post-trained with contrastive or instruction-tuned sentence objectives: BGE-base-en-v1.5~\citep{xiao2024c}, MPNet-base sentence encoder~\citep{reimers2019sentence,song2020mpnet}, Sentence-T5-base~\citep{ni2022sentence}, and Instructor-base~\citep{su2023one}. 
\vspace{-0.5em}

\paragraph{Encoder--decoders.}
Four sequence-to-sequence models: T5-v1.1-base~\citep{raffel2020exploring}, Flan-T5-base~\citep{chung2024scaling}, ByT5-base~\citep{xue2022byt5}, and BART-base~\citep{lewis2020bart}. 
\vspace{-0.5em}

\paragraph{Decoder-only models.}
Four autoregressive decoders, evaluated only on multiple-choice QA (Section~\ref{sec:tasks}) since left-to-right attention precludes symmetric pair encoding: GPT-2-base~\citep{radford2019language}, GPT-Neo-125M~\citep{black2021gpt}, Pythia-70M~\citep{biderman2023pythia}, and Qwen2.5-0.5B~\citep{yang2024qwen2}. 
\vspace{-0.5em}

\paragraph{State space models.}
Mamba-130M~\citep{gu2023mamba} represents sub-quadratic sequence models without self-attention.
\vspace{-0.75em}

\subsection{Tasks and Datasets}
\label{sec:tasks}
\vspace{-0.5em}

We evaluate across four task categories and 13 datasets from MTEB~\citep{muennighoff2023mteb,enevoldsen2025mmteb} and related benchmarks (Table~\ref{tab:datasets}).
\vspace{-0.5em}

\paragraph{Classification.}
Four datasets covering single-sentence and intent classification: SST-2~\citep{socher2013recursive}, AG~News~\citep{zhang2015character}, Emotion~\citep{saravia2018carer}, and Banking77~\citep{casanueva2020efficient}. 
\vspace{-0.5em}

\paragraph{Pair Classification.}
Four datasets require comparison of two input sentences: SNLI~\citep{bowman2015large}, MultiNLI~\citep{williams2018broad}, MRPC~\citep{dolan2005automatically}, and QQP, part of GLUE dataset~\citep{wang2019glue}. Merging two representations for prediction makes pooling strategy consequential.
\vspace{-0.5em}

\paragraph{Semantic Textual Similarity.}
Graded similarity on STS-B~\citep{cer2017semeval} and BIOSSES~\citep{souganciouglu2017biosses} (biomedical).
\vspace{-0.5em}
\vspace{-1em}

\paragraph{Multiple-Choice QA.}
Three commonsense benchmarks: ARC~\citep{clark2018think}, HellaSwag~\citep{zellers2019hellaswag}, and PIQA~\citep{bisk2020piqa}.

All details of the probing setup are presented in Appendix~\ref{sec:probing}.

\subsection{Implementation Details}
\label{sec:impl}
\vspace{-0.5em}

All experiments are implemented in PyTorch and managed with Hydra for reproducible configuration. Probe training uses a batch size of 128; full probe and selector hyperparameters are listed in Appendix~\ref{app:hyperparameters}. Geometry metrics are computed on a 10\%-sampled subset of each layer's representations with a random seed of 42. All experiments were done on an A4500 NVIDIA GPU with 20GB VRAM.
\vspace{-0.75em}

\section{Layer-wise Analysis}
\vspace{-0.75em}

\label{sec:layerwise-analysis}
Using the results of our empirical study, we examine the raw layer-by-layer signal that SGI exploits: how probe performance evolves across depth for each architectural family, pooling, and task, and how the geometric quantities of Section~\ref{sec:methodology} track it. 

\begin{figure}[t]
\centering

\includegraphics[
    width=\linewidth,
    trim={0 0 3cm 0},
    clip
]{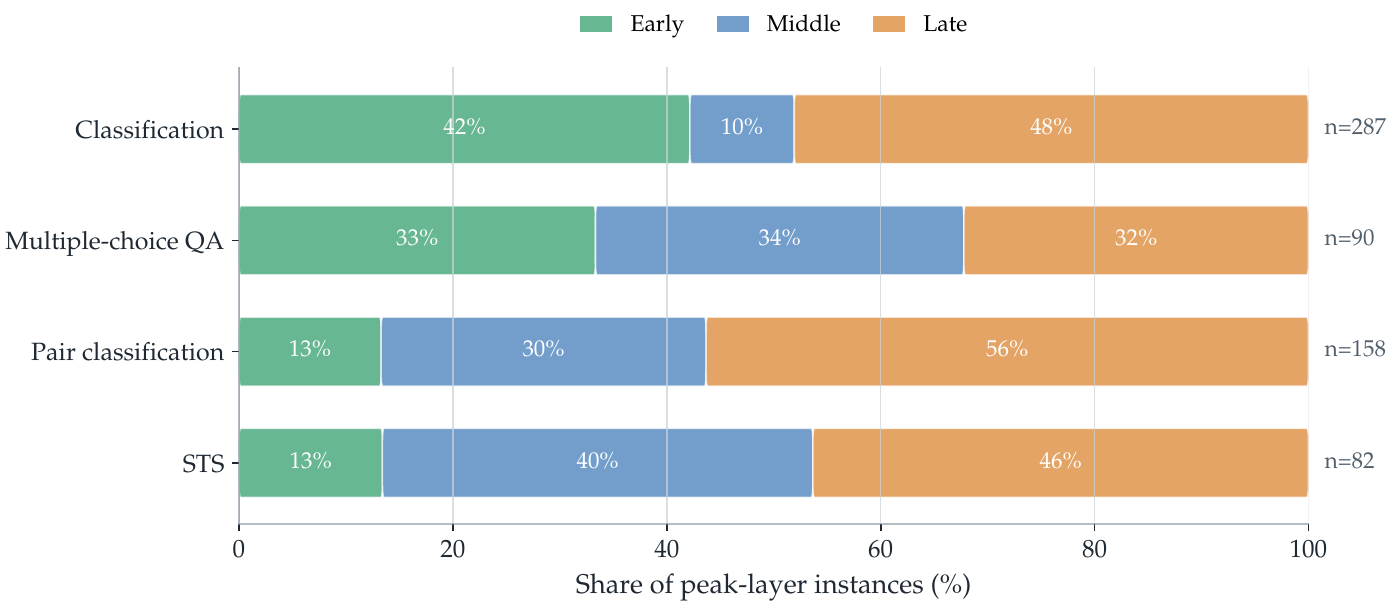}

\vspace{0.8em}

\includegraphics[
    width=\linewidth,
    trim={0 0 3cm 0},
    clip
]{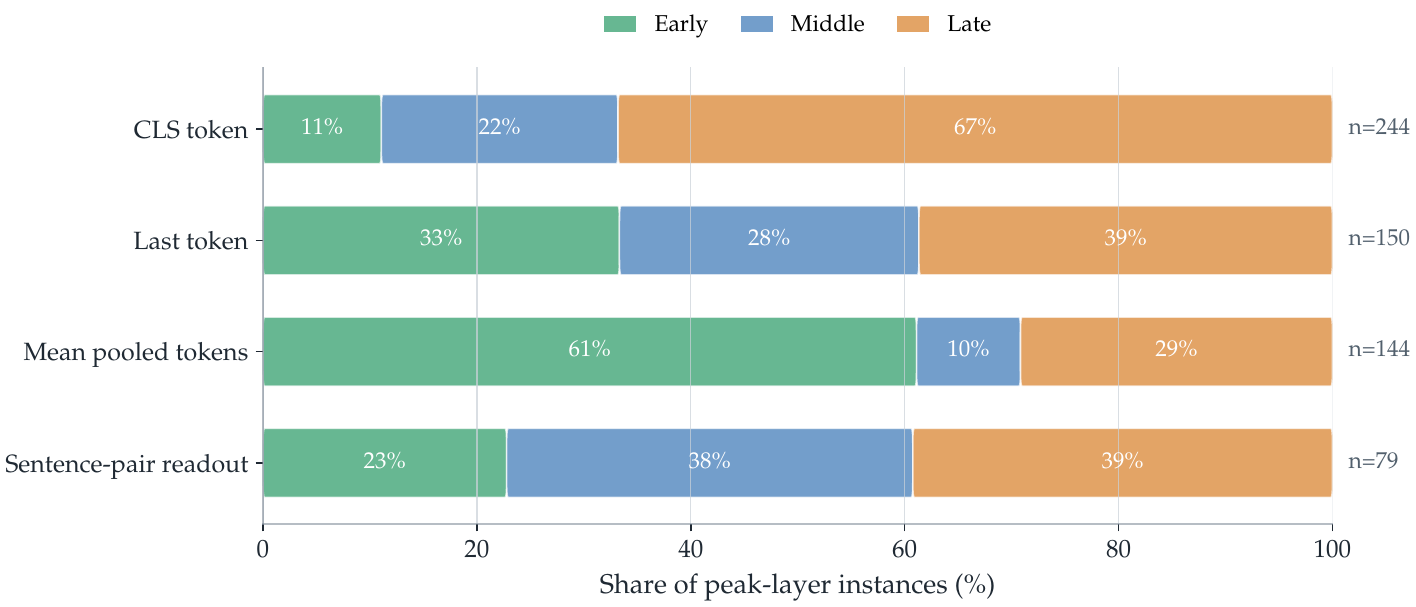}

\vspace{0.6em}

\caption{Empirical best-layer zones by task and readout over 4 tasks, 13 datasets and 25 models.}
\label{fig:layerwise-interpretability-main}
\vspace{-2em}

\end{figure}

Across most of the peak-layer instances, useful representations are not concentrated uniformly at the final layer.  Figure~\ref{fig:layerwise-interpretability-main} shows that the empirical best layer is organized by task and readout.  Classification has the strongest early component, with 42.2\% of runs peaking in the first third of the model, while pair classification and STS are more often middle or late.  Readout choice sharpens this pattern: mean pooling peaks early in 61.1\% of runs, whereas CLS readouts peak late in 66.8\% of runs.  

Mean pooling exposes early token-level evidence for classification, while CLS and sentence-pair readouts depend on later aggregation; multiple-choice QA spreads evenly across zones, consistent with answer scoring relying on endpoint representations. The implication for selection is that neither a fixed last-layer nor a fixed early-layer rule suffices: the target is a task- and readout-dependent quality phase, and because peak layers are structured rather than random, a fixed label-free score can track it.
Finer-grained analyses by architecture, pretraining objective, and near-best layer behavior are provided in Appendix~\ref{app:layer_analysis}.

\vspace{-0.5em}

\begin{figure*}[t]
\centering
\includegraphics[width=.8\textwidth]{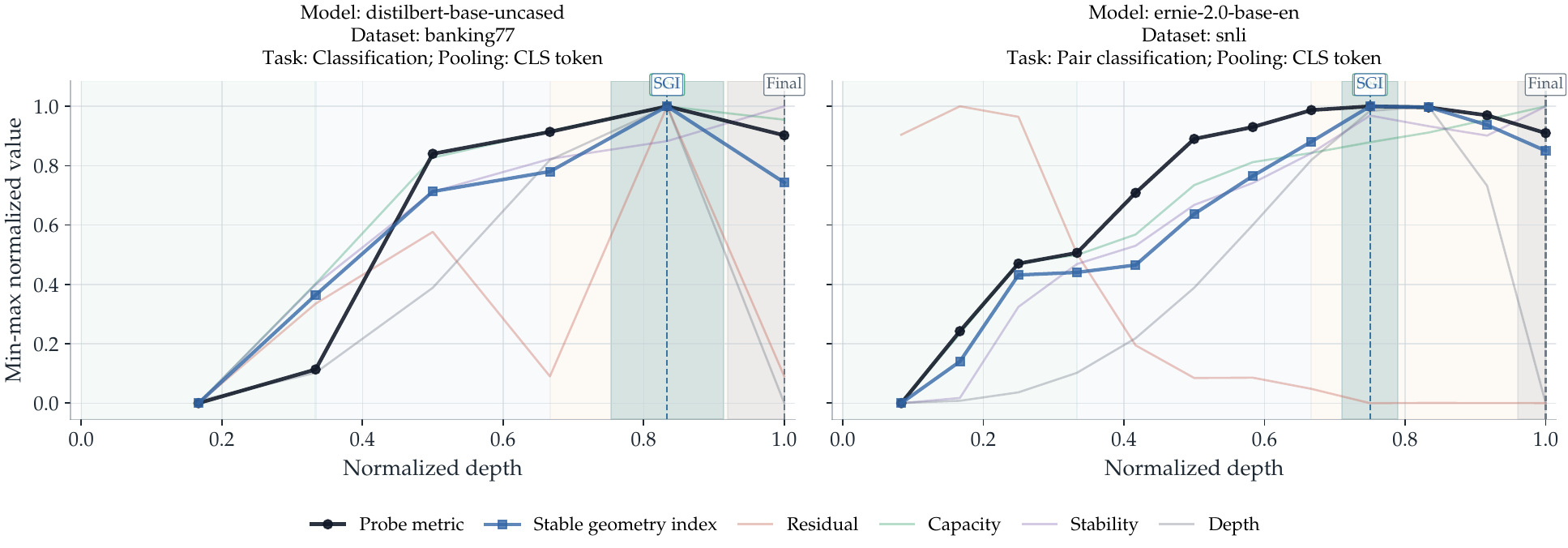}
\caption{Geometry-performance overlays.  Black denotes normalized probe performance, blue denotes SGI, and the dashed line marks the SGI-selected layer; in both examples, this layer is among the top three probe layers.}
\label{fig:app-overlay-positive}
\vspace{-1em}

\end{figure*}

\vspace{-0.5em}

\begin{figure}[h]
\centering
\includegraphics[width=\columnwidth]{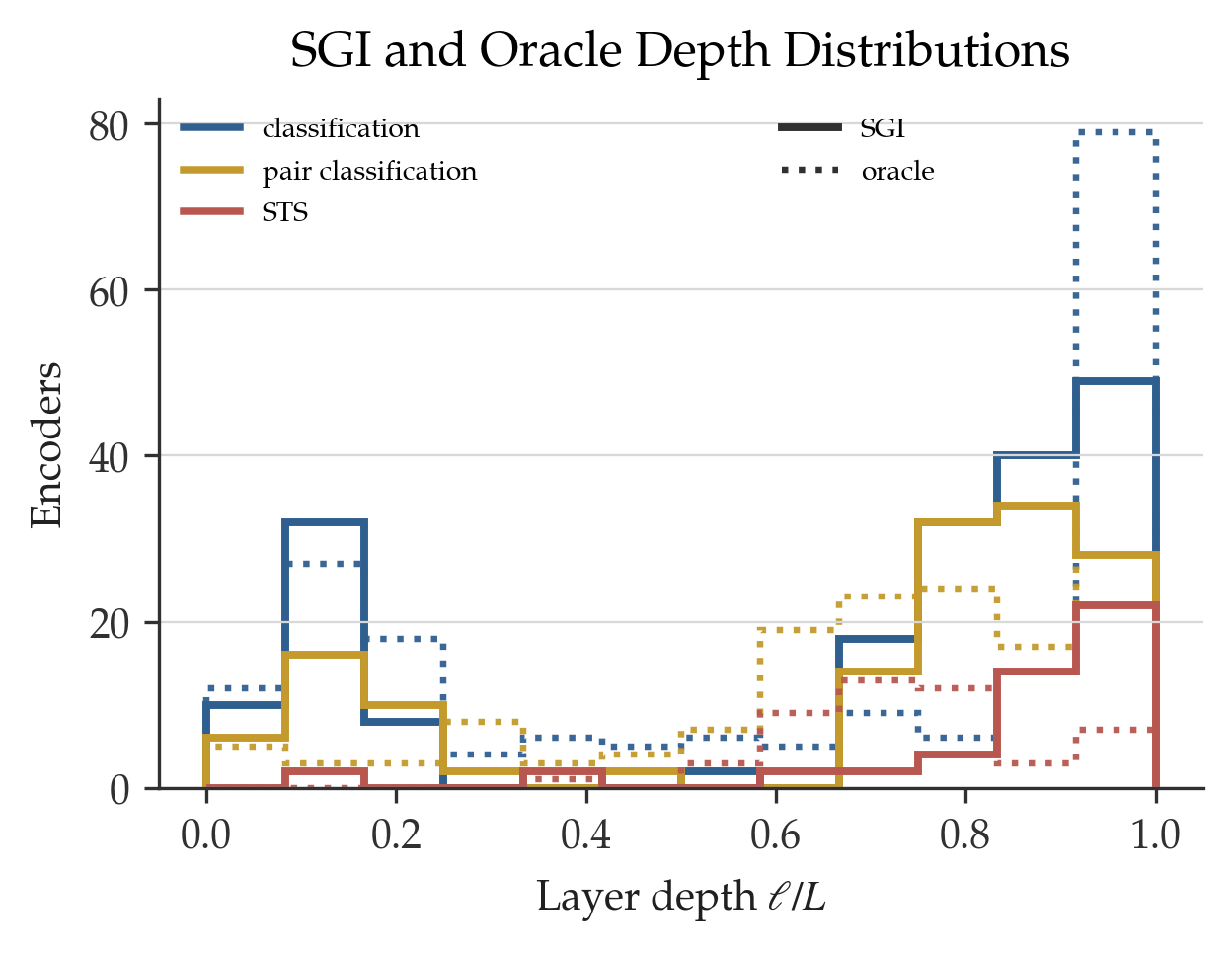}
\caption{Depth distributions of oracle layers (dotted) and layers selected by SGI (solid) across encoder tasks. SGI closely follows the oracle across all tasks, reproducing both the layer preference for and the overall distributional shape.}
\label{fig:SGI_oracle}
\vspace{-0.75em}

\end{figure}

\section{Results and Discussion}
\label{sec:results}
\vspace{-0.5em}

For a fixed (model, dataset, task, split) configuration with
candidate layers $1, \ldots, L$, a layer selector returns
$\hat\ell \in \{1, \ldots, L\}$. We report four metrics:
\begin{align*}
  \text{Gain-L\%} &= [M(\hat{\ell}) - M(L)] \times 100, \\
  \text{Gap-O (Oracle) \%} &= \![\max_j M(j) - M(\hat{\ell})] \times 100, \\
  \text{hit@1 (H@1)}       &= \indicator[\hat\ell \in \argmax_j M(j)], \\
  \text{hit@3 (H@3)}       &= \indicator[\hat\ell \in \Top_3],
\end{align*}

where \(M(\cdot)\) denotes the task-specific evaluation metric (see Table~\ref{tab:app-primary-metrics} for detailed mapping).

The central result is that this geometric test behaves like a practical selector, not a post-hoc description of representations (Figure~\ref{fig:app-overlay-positive}). Table~\ref{tab:overall} reports the main encoder evaluations. Table~\ref{tab:paired_stats} further confirms that these improvements are statistically significant over all 383 encoder evaluation groups. The primary metric improves by +1.20 percentage points (paired bootstrap 95\% CI [0.85, 1.54]; paired Wilcoxon, $p < 10^{-4}$), while Hit@3 improves by +12.79 percentage points (95\% CI [6.79, 18.80]; $p < 10^{-4}$). SGI improves over the final-layer baseline by $1.20$ percentage points and raises Hit@3 from $0.389$ to $0.517$.  The oracle improves over the final layer by $3.00$ percentage points. SGI not only recovers about $40\%$ of the available oracle gain without seeing probe labels, but also captures the layer selection pattern (Figure \ref{fig:SGI_oracle}) of the oracle layers.

\begin{table}[t]
\centering
\small
\setlength{\tabcolsep}{2pt}
\begin{tabular}{@{}lrrrr@{}}
\toprule
Method & Gain-L (\%) & Gap-O (\%) & H@1 & H@3 \\
\midrule
SGI (Our Method) & \textbf{1.20} & \textbf{1.80} & \textbf{0.285} & \textbf{0.517} \\
\hline
Eff. rank & 0.45 & 2.55 & 0.238 & 0.428 \\
RankMe & 0.45 & 2.55 & 0.238 & 0.428 \\
Last & 0.00 & 3.00 & 0.235 & 0.389 \\
Entropy & -0.29 & 3.28 & 0.198 & 0.397 \\
Part. ratio & -0.71 & 3.71 & 0.222 & 0.407 \\
Stable rank & -0.73 & 3.72 & 0.217 & 0.402 \\
Low top-1 var ratio & -0.73 & 3.72 & 0.217 & 0.402 \\
Low anis. & -0.88 & 3.88 & 0.214 & 0.394 \\
Low cond. & -0.90 & 3.89 & 0.196 & 0.371 \\
Intr. dim. & -0.90 & 3.89 & 0.154 & 0.347 \\
IsoScore & -2.07 & 5.06 & 0.159 & 0.352 \\
Low iso-ratio & -8.99 & 11.99 & 0.013 & 0.052 \\
\hline
\textit{Oracle} & 3.00 & 0.00 & 1.000 & 1.000 \\
\bottomrule
\end{tabular}
\caption{Encoder results.  Gain-L and Gap-O are in percentage points.}
\label{tab:overall}
\vspace{-1em}

\end{table}
\begin{table}[t]
\centering
\small
\setlength{\tabcolsep}{4pt}
\begin{tabular}{lccc}
\toprule
Metric & $\Delta$ & 95\% CI & Test \\
\midrule
Primary metric & +1.20 pp & [0.85, 1.54] & $p<10^{-4}$ \\
Hit@3 & +12.79 pp & [6.79, 18.80] & $p<10^{-4}$ \\
\bottomrule
\end{tabular}
\caption{Paired significance tests over the full 383-group encoder selector evaluation, with no model, dataset, or positive-gain filtering.}
\label{tab:paired_stats}
\end{table}

\begin{figure}[t]
\centering
\includegraphics[width=\columnwidth]{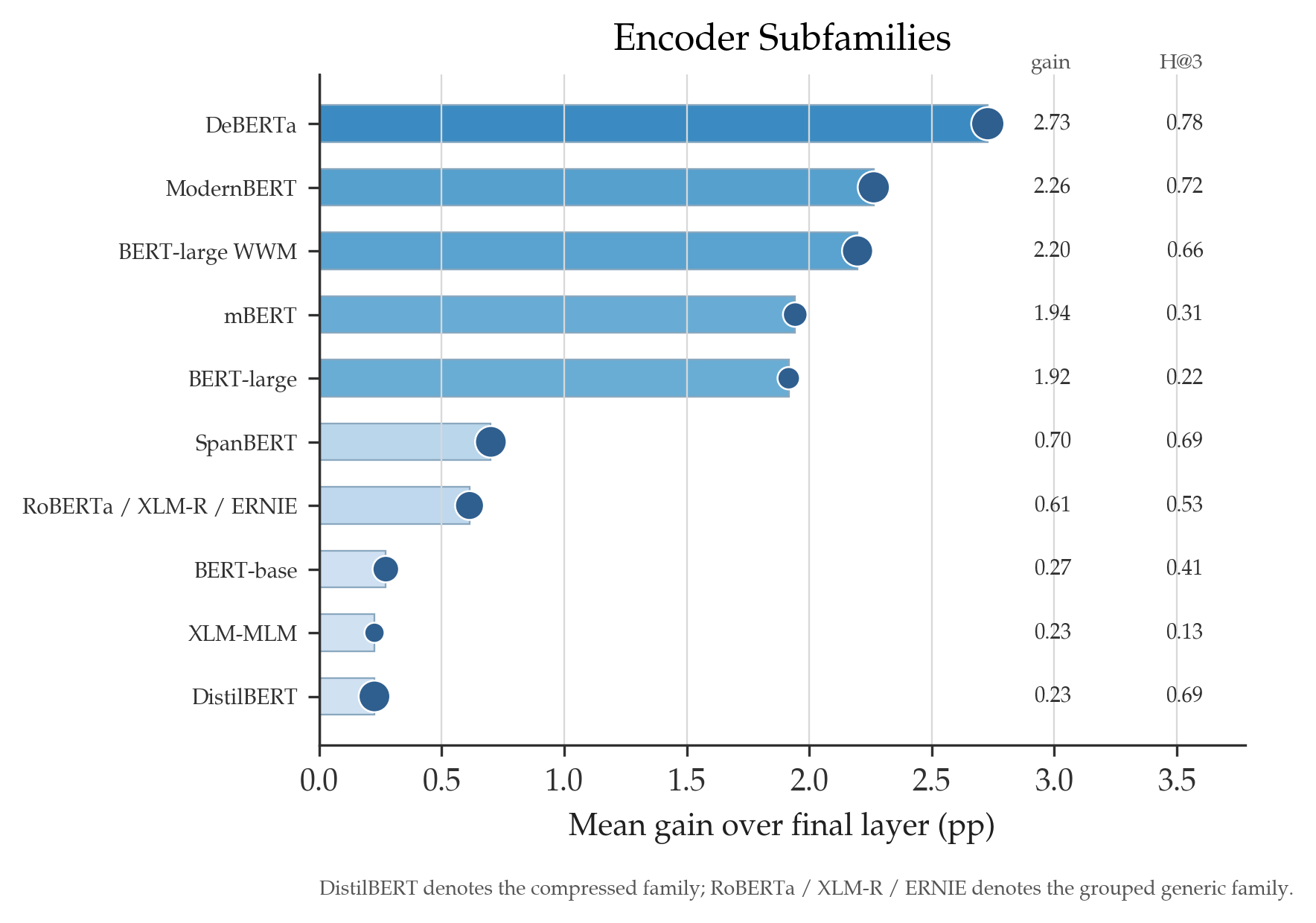}
\caption{Encoder-subfamily SGI gains. Bars show mean gain over the final layer
in percentage points; dots and right-hand values show Hit@3. SGI improves every
subfamily, with the largest gains for DeBERTa, ModernBERT, and BERT-large WWM.}
\label{fig:sgi-family-gain}
\end{figure}

The single-signal baselines show why a geometry index is needed.  Effective rank and RankMe are the strongest individual signals, but each reaches only $0.45$ pp gain.  Entropy, isotropy, intrinsic dimension, and spectral-floor selectors are useful diagnostics, yet none is a reliable selector by itself.

Table~\ref{tab:readout} partitions the encoder results across pooling strategies. Mean pooling benefits most in absolute gain, while CLS has the strongest Hit@3. This indicates that SGI is not merely avoiding one bad pooling choice; it remains useful across token-level, pooled, and pairwise pooling strategies.

SGI yields larger gains in DeBERTa, ModernBERT, and BERT-large variants (Figure~\ref{fig:sgi-family-gain}). These cross-model differences also vary with architecture and pretraining, so they do not isolate a causal effect of parameter scale. Section~\ref{subsec:scaling} provides a separate probe on larger checkpoints.

\begin{table}[t]
\centering
\scriptsize
\setlength{\tabcolsep}{3pt}
\begin{tabular}{lrrrr}
\toprule
Pooling & $n$ & Gain-L (pp) & Gap-O (pp) & H@3 \\
\midrule
CLS         & 191 & 0.58 & 1.80 & 0.613 \\
Last token  & 24  & 1.30 & 5.71 & 0.500 \\
Mean        & 96  & 2.48 & 1.13 & 0.427 \\
SBERT pair  & 72  & 1.09 & 1.40 & 0.389 \\
\bottomrule
\vspace{-1em}

\end{tabular}
\caption{Per-pooling SGI performance on encoders. Mean pooling gives the largest gain
over the final layer, and CLS gives the highest Hit@3.}
\label{tab:readout}

\vspace{-1em}

\end{table}
\vspace{-0.75em}

\subsection{Beyond Encoders}
\label{subsec:beyond-encoders}
\vspace{-0.5em}

A broader model-family analysis asks whether the same geometric principle continues to behave sensibly outside the encoder setting.  Table \ref{tab:model-family} shows a clear pattern: SGI is strongest when intermediate representations are expected to reorganize substantially across depth, especially for encoder--decoder and sentence-transformer models.  Decoder-only models show smaller gains, suggesting that the final layer is already a better default for these settings. Across tasks, classification benefits most, STS remains positive but farther from the oracle, and multiple choice QA is close to the final-layer baseline, indicating that this task family may require signals beyond layer geometry alone (Table \ref{tab:all-task}).

The 383-group encoder evaluation is our best-supported result. The decoder-only gain is only $+0.11$ pp, and the state-space result covers one model across eight groups; the latter is limited evidence rather than a family-level conclusion.

\begin{table}[]
\centering
\scriptsize
\setlength{\tabcolsep}{3pt}
\begin{tabular}{lrrrr}
\toprule
Task & $n$ & Gain-L (\%) & Gap-O (\%) & H@3 \\
\midrule
Classification & 287 & 3.18 & 1.97 & 0.585 \\
Multiple-choice QA & 90 & 0.06 & 1.10 & 0.378 \\
Pair classification & 158 & 0.76 & 1.39 & 0.399 \\
STS & 82 & 1.69 & 9.41 & 0.476 \\
\bottomrule
\end{tabular}
\caption{Task-wise behavior across model families.  SGI gives the largest task-level gain on classification, remains positive on STS and pair classification, and is close to the final-layer baseline on multiple-choice QA.}
\label{tab:all-task}
\vspace{-1em}

\end{table}
\vspace{-0.5em}

\begin{table}[]
\centering
\scriptsize
\setlength{\tabcolsep}{4pt}
\begin{tabular}{lrr}
\toprule
Method & Metric & Gain-L (\%) \\
\midrule
Oracle best layer & 0.715 & 2.38 \\
SGI & 0.697 & 0.58 \\
Last layer & 0.692 & 0.00 \\
Learned all-layer mix & 0.683 & -0.83 \\
\bottomrule
\end{tabular}
\caption{SGI versus a supervised all-layer mix on the matched CLS token ($n=191$).  SGI exceeds the learned mix by $1.41$ metric points and exceeds the final layer by $0.58$ pp, despite using no labels for layer selection.}
\label{tab:learned}
\end{table}

\begin{table}[t]
\centering
\scriptsize
\setlength{\tabcolsep}{3pt}
\begin{tabular}{lrrrr}
\toprule
Model family & $n$ & Gain-L (\%) & Gap-O (\%) & H@3 \\
\midrule
Encoder & 383 & 1.20 & 1.80 & 0.517 \\
Decoder & 48 & 0.11 & 1.49 & 0.354 \\
Encoder--decoder & 80 & 6.51 & 5.91 & 0.600 \\
Sentence transformer & 82 & 2.33 & 2.19 & 0.598 \\
State-space & 8 & 4.57 & 1.50 & 0.125 \\
\bottomrule
\end{tabular}
\caption{Model-family behavior.  SGI is strongest for encoder--decoder and sentence-transformer models, positive for encoders and state-space models, and smallest for decoder-only models.}
\label{tab:model-family}
\vspace{-1em}

\end{table}

\subsection{Larger-model Scaling Probe}
\label{subsec:scaling}

To test whether the pipeline extends beyond the original cohort, Table~\ref{tab:large-model-scaling} evaluates four additional checkpoints. Across 57 paired evaluation rows per method, SGI has the highest Hit@1 among the compared label-free selectors and ties for the highest Hit@3. Its mean gain over the final layer is positive for every checkpoint: $+5.11$ pp for DeBERTa-v3-large, $+4.96$ pp for Mamba-2.8B, $+2.78$ pp for XLM-RoBERTa-large, and $+0.89$ pp for Qwen3.5-4B.

\begin{table}[t]
\centering
\footnotesize
\setlength{\tabcolsep}{3.5pt}
\begin{tabular}{@{}llr@{}}
\toprule
Model & Datasets & Gain-L (pp) \\
\midrule
DeBERTa-v3-large   & SST-2, MRPC, STS-B & $+5.11$ \\
Mamba-2.8B         & ARC, PIQA, STS-B   & $+4.96$ \\
XLM-RoBERTa-large  & SST-2, MRPC, STS-B & $+2.78$ \\
Qwen3.5-4B         & ARC, PIQA          & $+0.89$ \\
\bottomrule
\end{tabular}
\caption{\textbf{Larger-model scaling probe.} Mean gain over the final layer in percentage points. The probe contains 57 paired rows per method across four checkpoints, including six for Qwen3.5-4B, whose mean remaining oracle gap is $2.90$ pp.}
\label{tab:large-model-scaling}
\end{table}

This is evidence that SGI can retain useful selection signal on larger checkpoints, rather than a controlled scaling law or broad validation at billion-parameter scale. The 57 scaling rows are separate from the original 383-group encoder result and its significance tests.

\subsection{Comparison to Supervised Layer Mixing}
\label{subsec:supervised-mix}
\vspace{-0.25em}

We compare SGI against a supervised all-layer scalar mix (Table \ref{tab:learned}), following the ELMo-style layer-weighting paradigm \citep{peters2018deepcontextualizedwordrepresentations}.  The scalar mix learns layer weights from probe-supervised training data, whereas SGI selects a single layer from unlabeled geometry.

\begin{table}[t]
\centering
\scriptsize
\setlength{\tabcolsep}{4pt}
\begin{tabular}{lrrr}
\toprule
Task & $n$ & SGI & Learned \\
\midrule
Classification & 95 & 0.740 & 0.720 \\
Pair classification & 72 & 0.655 & 0.656 \\
STS & 24 & 0.657 & 0.620 \\
\bottomrule
\end{tabular}
\caption{Task-wise supervised comparison.  SGI is higher on classification and
STS, and nearly tied on pair classification.}
\label{tab:learned-task}
\vspace{-1em}

\end{table}

\vspace{-0.5em}
\vspace{-0.5em}

\subsection{Ablations}
\vspace{-0.5em}

\label{subsec:ablations}
\begin{table}[t]
\centering
\scriptsize
\setlength{\tabcolsep}{4pt}
\begin{tabular}{@{}llrr@{}}
\toprule
Variant & Score & Gain-L (\%) & H@3 \\
\midrule
SGI (full)          & $R+C+S+P$         & \textbf{1.20} & \textbf{0.517} \\
\midrule
No residual         & $C+S+P$           & 0.72          & 0.473 \\
No capacity         & $R+S+P$           & 0.00          & 0.454 \\
No spread           & $R+C+P$           & 0.94          & 0.478 \\
No depth prior      & $R+C+S$           & 0.73          & 0.449 \\
\midrule
Linear capacity     & $R+Z+S+P$         & 0.96          & 0.491 \\
Product interaction & $R+C+SP$          & $-0.17$       & 0.420 \\
\midrule
Residual $\times2$  & $2R+C+S+P$        & 1.08          & 0.499 \\
Capacity $\times2$  & $R+2C+S+P$        & 0.93          & 0.478 \\
Stability $\times2$ & $R+C+2S+P$        & 0.89          & 0.499 \\
Depth $\times2$     & $R+C+S+2P$        & 0.38          & 0.441 \\
\bottomrule
\end{tabular}
\caption{Ablations of the SGI score on 383 encoder evaluation groups.
The second block removes one term at a time. The third block replaces
the capacity parameterisation with its linear form ($Z_\ell=\Lup(e)_\ell$
without the sharpening factor $(1{+}Z_\ell)$) and replaces the additive
stability--depth combination with a multiplicative interaction. The final
block doubles the weight of each SGI component individually while keeping
the remaining terms unchanged. Equal-weight SGI remains the best-performing
variant.}
\label{tab:ablation}
\end{table}

Table~\ref{tab:ablation} probes both \emph{which terms} SGI requires
and \emph{which functional form} it should take. The middle block shows
that all four terms carry non-redundant signal: removing any one reduces
both Gain-L and Hit@3. Capacity is the most consequential term, and its
removal collapses the selector to the last-layer baseline, confirming
that spectral breadth is necessary for any layer to be preferred over
the final layer, but insufficient to determine layer quality. Residual,
spread, and depth contribute smaller but reliable margins of 0.48, 0.26,
and 0.47 pp respectively, together converting broad rank into a stable
selection signal.

The lower block studies score formulation and weighting. Capacity is parameterised as
$C_\ell = Z_\ell(1+Z_\ell)$ rather than linear $Z_\ell$; this is not
cosmetic, as the linear form loses 0.24 pp in Gain-L, showing that
sharpened response near high effective rank carries information beyond
effective rank alone. We also evaluate weighted variants by doubling each SGI component individually while keeping the remaining terms fixed. None of these variants exceeds the equal-weight formulation. This is a coarse sensitivity check, not a full weight search, so it supports the fixed formulation among the tested variants without establishing globally optimal weights. More importantly, additivity is not
interchangeable with multiplicative composition. Replacing the
stability--depth sum with a product $S \cdot P$ drives Gain-L below
zero, making the selector worse than the last-layer default. This
failure is structural: in a product, one large signal can mask weakness
in another, so a layer with broad capacity and favourable depth can win
despite poor stability. The additive form avoids this particular interaction. The observed results across tasks, readouts, and architectures (Tables~\ref{tab:model-family}, \ref{tab:all-task}) provide empirical support for the combination; neither these results nor the ablations establish conditional independence or robustness to arbitrary estimation errors.

\begin{figure}[t]
\centering
\includegraphics[width=\columnwidth]{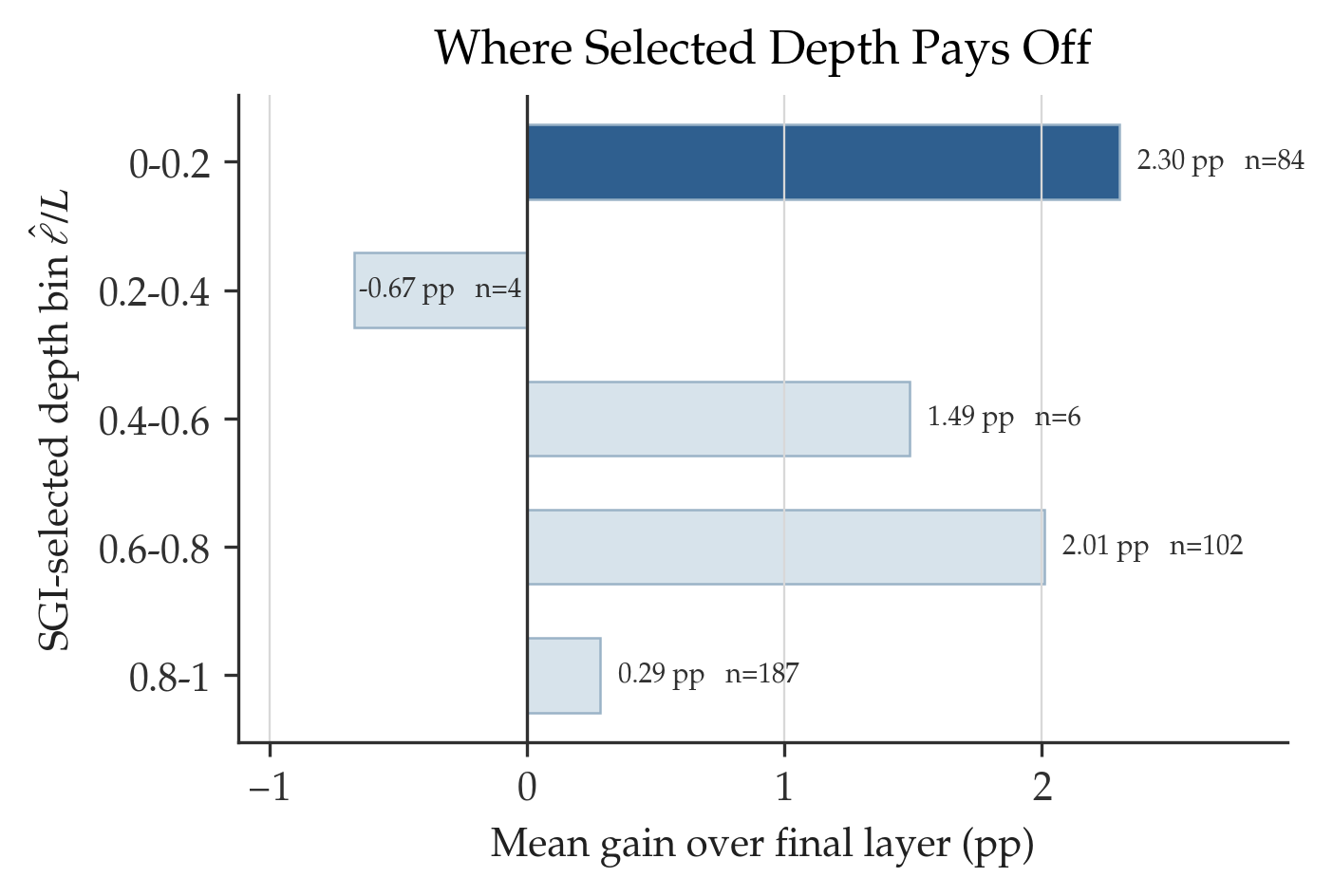}
\caption{Mean gain of SGI over the last layer, binned by the normalised
depth $\hat\ell/L$ of the SGI-selected layer. The shallowest bin
($\hat\ell/L < 0.2$) carries the largest mean gain ($2.30$ pp, $n=84$),
identifying settings where the upper $80\%$ of an encoder's layers can
be pruned without loss of probe quality. The deepest bin ($\hat\ell/L
\geq 0.8$, $n=187$) shows a smaller gain, consistent with the final
layer being already close to the optimum in those settings.}
\label{fig:depth-gain}
\vspace{-1.5em}

\end{figure}

\vspace{-0.5em}

\paragraph{Practical implications.}
SGI changes two practitioner workflows. For \emph{probing}, it removes the need to train a probe at every layer before choosing one, since the selection score is computed from unlabeled representations, applies the same rule across task families and readouts, and can serve as a strong default layer or initialisation point when labels become available later. For \emph{deployment}, the depth distribution of SGI-selected layers carries direct consequences for model truncation and layer pruning: Figure~\ref{fig:depth-gain} bins the 383 encoder evaluation groups by the normalised depth $\hat\ell/L$ of the SGI-selected layer and shows that the shallowest bin ($\hat\ell/L < 0.2$, 84 groups, or $22\%$ of encoder workload) carries the largest mean gain over the final layer at 2.30 pp, identifying settings where the upper $80\%$ of an encoder's layers can be removed without loss of probe quality. For a 12-layer encoder selected at $\hat\ell = 2$, this corresponds to a $5\times$ reduction in inference FLOPs and parameter footprint at deployment time, while improving rather than degrading the downstream probe metric. The early-selection bin is not an artifact of the depth prior, since $P_\ell \propto d_\ell^4(1-d_\ell)$ places its mode at $d=0.8$ and would otherwise push selections deeper; early selections occur only when the spectral terms ($R$, $C$, $S$) collectively outvote the prior, making them evidence-driven cutoff signals rather than depth-prior side effects. Unlike learned-layer-mixing approaches \citep{peters2018deepcontextualizedwordrepresentations} that require probe labels to identify removable layers, SGI provides this signal from unlabeled forward passes alone, making it applicable in pre-label model selection, in zero-resource deployment pipelines, and as a label-free initialisation for static pruning and knowledge-distillation procedures that need to choose a truncation depth before any downstream task is fixed.

\vspace{-0.75em}
\section{Conclusion}
\label{sec:conclusion}
\vspace{-0.75em}

We asked where representation quality lives in pretrained text models
and whether that location is recoverable from unlabeled geometry. Peak
layers are not concentrated at the endpoint but organised by the
interaction of objective, task, and readout, so no fixed depth rule
transfers. The Stable Geometry Index exploits this: a closed-form, fixed-form
score with no fitted coefficients that improves over the final-layer baseline,
recovers about $40\%$ of the oracle gap without labels, and exceeds a
supervised layer mix on the matched subset, while its ablations show
power-law fit quality, not the exponent, carries the spectral signal.
Needing only unlabeled forward passes, it also serves as a label-free
pruning prior. SGI is best read as a phase detector rather than a
universal proxy: strongest where representations reorganise across
depth, and weakest on multiple-choice QA where the final layer is
already competitive.

\section{Limitations}
SGI is weakest on multiple-choice  QA, where the final layer is already competitive and the oracle gap is small; we expect that tasks requiring stronger generative or retrieval signals may benefit  from geometry measures beyond those considered here, and extending the framework to generation and structured prediction settings is a direction we leave for future work. SGI's additive form is empirically supported on the evaluated settings, but its terms need not be statistically independent, and geometry alone does not guarantee downstream label separability. Finally, SGI 
selects a single layer, and combining it with lightweight layer mixing, using 
SGI as an initialisation rather than a hard selector, could close the remaining 
gap to the oracle while retaining the label-free property that makes it practical 
for pre-label model selection and knowledge distillation pipelines.

\section{Ethical Considerations}
This work studies layer selection and representation geometry for pretrained language models. It uses only publicly available checkpoints and benchmark datasets, and does not involve human subjects, private data, or any new data collection. The method is intended for offline research analysis and efficiency improvements in inference, not for profiling individuals or making decisions about people. By identifying layers that can be skipped without hurting performance, the approach may reduce compute, latency, and energy use. At the same time, the method does not remove the broader risks associated with the underlying models, and any deployment should still follow the license and use restrictions of the original checkpoints and datasets.

\appendix
\clearpage
\section{Appendix}
\label{sec:appendix}

This appendix provides the supporting evidence behind the main paper. Appendices~\ref{sec:probing} expands the probing setup from Section~\ref{sec:experiments}, and Appendix~\ref{sec:interp-appendix} supports the two central empirical claims of Section~\ref{sec:layerwise-analysis}: (a) that useful representation layers are structured by task, readout, architecture family, and pretraining objective rather than being uniformly concentrated at the final layer; and (b) that the four geometric coordinates used by SGI, namely spectral capacity, power-law fit quality, condition number, and local intrinsic dimensionality, each capture a distinct and non-redundant property of the representation manifold that correlates with probe-layer quality.

Appendix~\ref{sec:interp-appendix} is organised as follows. We first report the peak-zone tables behind Figure~\ref{fig:layerwise-interpretability-main} together with model-family, training-label, and readout stratifications. We then enumerate the observed model-family/training-label/task/readout combinations, separate where a readout peaks from which readout achieves the best score, and document the raw SGI coordinates and their signed differences relative to the final layer. The overlay figures provide qualitative alignment checks between the geometry curve and the supervised probe curve. Together these materials show that the structure reported in the main text holds across the full set of evaluated configurations.

\section{Probing Setup}
\label{sec:probing}
\vspace{-0.5em}

For each model and dataset we extract intermediate representations from all transformer layers (including the embedding layer), yielding a layer-by-layer profile per model--task pair.
\vspace{-0.5em}

\paragraph{Pooling.}
We reduce each token-level sequence to a fixed-size vector before probing using up to four strategies: \textsc{cls}-token (BERT-style encoders), \textsc{last-token} (decoder and encoder--decoder models), \textsc{mean} pooling across non-padding tokens, and \textsc{sbert-pair} (mean-of-pair with mean--max fusion for pair tasks).
\vspace{-0.5em}

\paragraph{Probe architectures.}
A linear probe (logistic regression) and a nonlinear probe (two-layer MLP with GeLU activations, hidden dimension 256, dropout $p{=}0.1$) are trained per layer. All probes use AdamW ($\mathrm{lr}{=}10^{-3}$, weight decay 0.01, batch size 128) for up to 20 epochs, with early-stopping patience 3, and report averaged results across three seeds.

\section{Layerwise Interpretability Details}
\label{sec:interp-appendix}

\subsection{Experimental Coverage and Evaluation Denominators}
\label{sec:app-coverage}

This subsection records the evaluation denominators used by the main analyses.
All peak-layer and selector quantities are computed using the task's primary
metric, denoted $M_g(\ell)$ for evaluation group $g$ at layer $\ell$.
Table~\ref{tab:app-primary-metrics} lists the metric used for each task family.
The empirical best layer, earliest near-best layer, gap-to-last, gap-to-best,
and Hit@1/Hit@3 oracle ranking are all defined with respect to this primary
metric. Thus comparisons across layers are task-internal: accuracy is used for
classification-style tasks, while Spearman correlation is used for STS.

\begin{table*}[t]
\centering
\small
\begin{tabular}{lll}
\toprule
Task family & Datasets & Primary metric \\
\midrule
Classification & SST-2; AG News; Emotion; Banking77 & Accuracy \\
Pair classification & SNLI; MultiNLI; MRPC; QQP & Accuracy \\
STS & STS-B; BIOSSES & Spearman correlation \\
Multiple-choice QA & ARC; HellaSwag; PIQA & Accuracy \\
\bottomrule
\end{tabular}
\caption{Primary evaluation metric used to define empirical best layers and
selector outcomes for each task family.}
\label{tab:app-primary-metrics}
\end{table*}

Table~\ref{tab:app-cache-scope} summarises the scope of the three analyses in the paper: the broad peak-layer analysis, the layerwise score set from which those peaks are computed, and the encoder-only SGI selector evaluation. The layerwise study spans the full model set, while the SGI selector results are reported on the encoder subset for which the selector is claimed.

\begin{table}[t]
\centering
\small
\begin{tabular}{llr}
\hline
\textbf{Dataset} & \textbf{Task} & \textbf{Domain} \\
\hline
SST-2     & classification      & Sentiment   \\
AG~News   & classification      & News        \\
Emotion   & classification      & Social      \\
Banking77 & classification      & Finance     \\
\hline
SNLI      & pair classification & NLI         \\
MultiNLI  & pair classification & Multi-genre \\
QQP       & pair classification & Questions   \\
MRPC      & pair classification & Paraphrase  \\
\hline
STS-B     & STS                 & General     \\
BIOSSES   & STS                 & Biomedical  \\
\hline
ARC       & MC-QA               & Science     \\
HellaSwag & MC-QA               & Commonsense \\
PIQA      & MC-QA               & Physical    \\
\hline
\end{tabular}
\caption{The 13 datasets used in our evaluation (STS\,=\,semantic textual similarity, MC-QA\,=\,multiple-choice QA).}
\label{tab:datasets}
\end{table}
\begin{table*}[t]
\centering
\small
\begin{tabular}{lrrr}
\toprule
Artifact & Rows & Models & Datasets \\
\midrule
Peak-layer instances & 617 & 25 & 13 \\
Layerwise score rows & 8,544 & 25 & 13 \\
Final SGI selector groups & 383 & 12 & 8 \\
\bottomrule
\end{tabular}
\caption{Scope of the three analyses in the paper. Peak-layer instances are computed after averaging probe scores across seeds.}
\label{tab:app-cache-scope}
\end{table*}

Table~\ref{tab:app-coverage-matrix} gives the configuration coverage for each analysis. A peak-layer instance is one seed-averaged (model, dataset, task, readout, probe) curve over all observed layers. The selector comparison evaluates 21 label-free methods on the same 383 encoder evaluation groups used for the main SGI result.

\begin{table*}[t]
\centering
\scriptsize
\resizebox{\textwidth}{!}{%
\begin{tabular}{lrrrrrrl}
\toprule
Analysis scope & Models & Datasets & Tasks & Readouts & Probes & Seeds & Evaluation unit \\
\midrule
Layerwise peak analysis & 25 & 13 & 4 & 4 & 2 & 3 & 617 peak-layer instances \\
Layerwise scores & 25 & 13 & 4 & 4 & 2 & 3 & 8,544 layerwise score rows \\
Final SGI selector evaluation & 12 & 8 & 3 & 4 & 2 & 3 & 383 encoder selector groups \\
Selector comparison & 12 & 8 & 3 & 4 & 2 & 3 & 21 methods on the same 383 groups \\
\bottomrule
\end{tabular}
}
\caption{Configuration coverage for the main empirical claims. The broad
layerwise analysis covers classification, pair classification, STS, and
multiple-choice QA. The final SGI selector evaluation is the encoder-only
subset used for the main selector results.}
\label{tab:app-coverage-matrix}
\end{table*}

Table~\ref{tab:app-task-readout-probe-support} breaks the 617 peak-layer instances down by task, readout, and probe type, showing which task--readout--probe combinations contribute to the reported early, middle, and late peak statistics.

\begin{table*}[t]
\centering
\scriptsize
\resizebox{\textwidth}{!}{%
\begin{tabular}{llrrr}
\toprule
Task & Readout & Linear probe & Nonlinear probe & Total \\
\midrule
Classification & CLS token & 54 & 53 & 107 \\
Classification & Last token & 18 & 18 & 36 \\
Classification & Mean pooled tokens & 72 & 72 & 144 \\
Multiple-choice QA & CLS token & 12 & 12 & 24 \\
Multiple-choice QA & Last token & 33 & 33 & 66 \\
Pair classification & CLS token & 40 & 39 & 79 \\
Pair classification & Sentence-pair readout & 40 & 39 & 79 \\
STS & CLS token & 17 & 17 & 34 \\
STS & Last token & 24 & 24 & 48 \\
\midrule
Total & -- & 310 & 307 & 617 \\
\bottomrule
\end{tabular}
}
\caption{Task--readout--probe support for the layerwise peak analysis. Counts
are peak-layer instances after seed averaging.}
\label{tab:app-task-readout-probe-support}
\end{table*}

Table~\ref{tab:app-sgi-selector-coverage} reports the coverage of the final SGI selector evaluation. The 383 groups span three encoder task families, four readouts, two probe types, and two evaluation-split regimes.

\begin{table}[t]
\centering
\small
\begin{tabular}{llr}
\toprule
Dimension & Value & Groups \\
\midrule
Task & Classification & 191 \\
Task & Pair classification & 144 \\
Task & STS & 48 \\
\midrule
Readout & CLS token & 191 \\
Readout & Mean pooled tokens & 96 \\
Readout & Sentence-pair readout & 72 \\
Readout & Last token & 24 \\
\midrule
Probe & Linear and nonlinear & 383 \\
\midrule
Evaluation split & Official test & 287 \\
Evaluation split & Random 60/20/20 & 96 \\
\midrule
Embedding split & Train & 383 \\
\bottomrule
\end{tabular}
\caption{Coverage of the final SGI selector evaluation. All rows are encoder
groups.}
\label{tab:app-sgi-selector-coverage}
\end{table}

\subsection{Peak-Zone Tables}
\label{subsec:layer-zones}
Tables~\ref{tab:app-task-zones} and~\ref{tab:app-readout-zones} provide the data behind the top panels of Figure~\ref{fig:layerwise-interpretability-main}.  The early, middle, and late zones correspond to the first, second, and final thirds of the model depth after normalizing by the final observed layer.

\begin{table}[h]
\centering
\scriptsize
\resizebox{\columnwidth}{!}{%
\begin{tabular}{lrrrr}
\toprule
Task & Early & Middle & Late & Total \\
\midrule
Classification & 121 (42.2\%) & 28 (9.8\%) & 138 (48.1\%) & 287 \\
Multiple-choice QA & 30 (33.3\%) & 31 (34.4\%) & 29 (32.2\%) & 90 \\
Pair classification & 21 (13.3\%) & 48 (30.4\%) & 89 (56.3\%) & 158 \\
STS & 11 (13.4\%) & 33 (40.2\%) & 38 (46.3\%) & 82 \\
\bottomrule
\end{tabular}
}
\caption{Best-layer zone counts by task.  Parentheses show percentages within task.}
\label{tab:app-task-zones}
\end{table}

\begin{table}[t]
\centering
\scriptsize
\resizebox{\columnwidth}{!}{%
\begin{tabular}{lrrrr}
\toprule
Readout & Early & Middle & Late & Total \\
\midrule
CLS token & 27 (11.1\%) & 54 (22.1\%) & 163 (66.8\%) & 244 \\
Last token & 50 (33.3\%) & 42 (28.0\%) & 58 (38.7\%) & 150 \\
Mean pooled tokens & 88 (61.1\%) & 14 (9.7\%) & 42 (29.2\%) & 144 \\
Sentence-pair readout & 18 (22.8\%) & 30 (38.0\%) & 31 (39.2\%) & 79 \\
\bottomrule
\end{tabular}
}
\caption{Best-layer zone counts by readout location.  Parentheses show percentages within readout.}
\label{tab:app-readout-zones}
\end{table}

\subsection{Detailed Layer-wise Analysis}
\label{app:layer_analysis}
Median peak depth rises from encoder--decoders ($0.167$) through decoders ($0.542$) to encoders and sentence transformers ($0.727$, $0.917$; Table~\ref{tab:app-family-quartiles}), encoder--decoders span an interquartile range of $0.750$. The finer strata (Tables~\ref{tab:app-main-strata}, \ref{tab:app-early-heatmap-data}) show the spread is driven by objective--task--readout interaction: span-corruption models are near-best extremely early on classification and STS ($100\%$), whereas masked language modeling is task-sensitive (early-accessible for $60.4\%$ of classification runs but only $3.1\%$ of STS). The useful layer is where a given objective and readout make the signal linearly available, not simply where the architecture peaks. A selector need not hit the exact argmax and may instead pick a near-best layer (within $0.01$ of the best score), so we also track these near-best layers.

This distinction matters when the exact optimum is hard to localize but a near-best region forms early. Causal decoders on multiple-choice QA are one such case: the exact best layer lies in the first third only $45.8\%$ of the time, yet an early layer reaches the near-best plateau in every such group (mean gap $0.005$; Tables~\ref{tab:app-main-strata}, \ref{tab:app-top-early}). Readout matters as much as objective: contrastive sentence models peak early under mean pooling ($91.7\%$, median depth $0.08$) but late under last-token readout (median depth $0.83$). 
\subsection{Model Family, Training Label, And Readout}

\begin{figure}[h]
\centering
\includegraphics[width=\columnwidth]{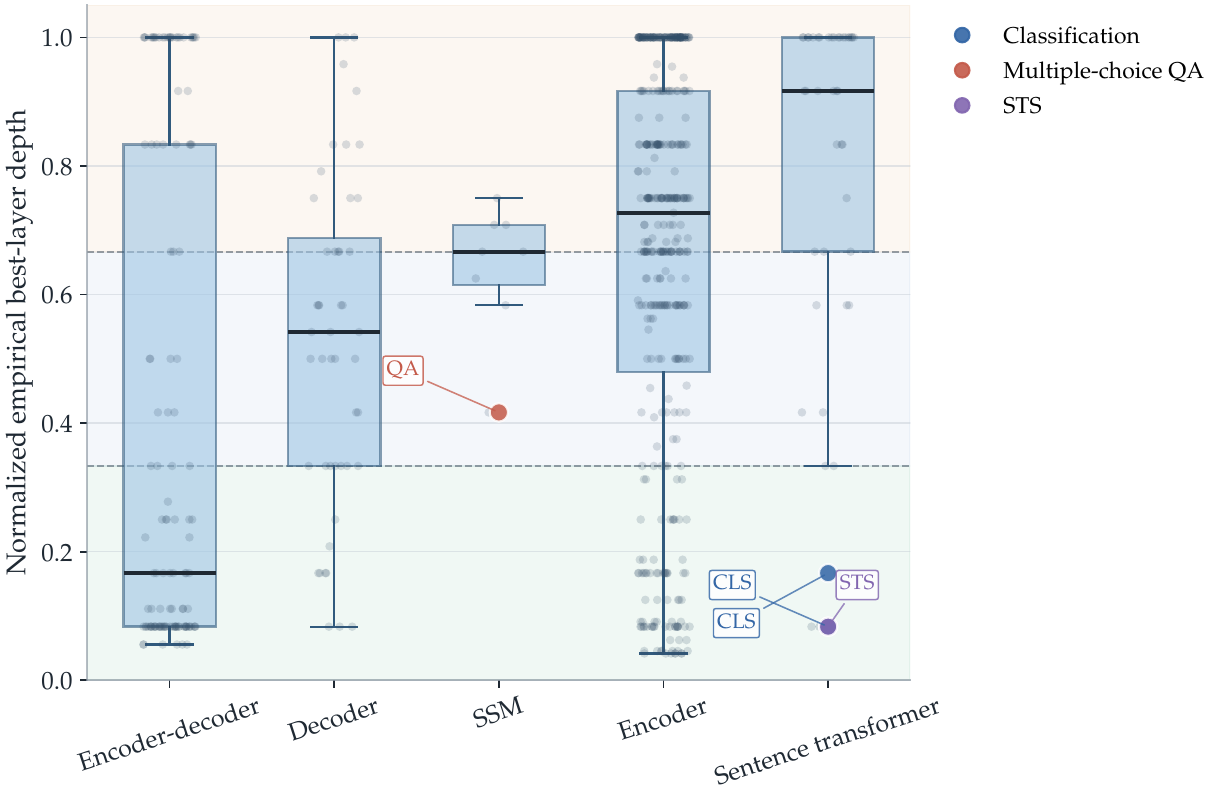}
\caption{Normalized best-layer positions by model family.  Outlier points are colored by task type and labeled with a short task tag.}
\label{fig:app-family-box}
\end{figure}

Table~\ref{tab:app-family-quartiles} summarizes normalized best-layer positions by model family.  Table~\ref{tab:app-main-strata} reports representative model-family, training-label, task, and readout strata, and Table~\ref{tab:app-top-early} lists the highest early near-best plateau groups. Figure~\ref{fig:app-family-box} shows the distribution of SGI scores across model families. The training labels are descriptive strata rather than causal explanations.  In particular, Contrastive Sentence Embedding Finetuning spans both encoder-only and encoder--decoder architectures; the label denotes the sentence-embedding finetuning stage rather than a specific base architecture.

\begin{table}[h]
\centering
\scriptsize
\resizebox{\columnwidth}{!}{\begin{tabular}{lrrrrrr}
\toprule
Family & N & Q1 & Median & Q3 & IQR & Outliers \\
\midrule
Encoder-decoder & 120 & 0.083 & 0.167 & 0.833 & 0.750 & 0 \\
Decoder & 48 & 0.333 & 0.542 & 0.688 & 0.354 & 0 \\
SSM & 8 & 0.615 & 0.667 & 0.708 & 0.094 & 1 \\
Encoder & 399 & 0.479 & 0.727 & 0.917 & 0.438 & 0 \\
Sentence transformer & 42 & 0.667 & 0.917 & 1.000 & 0.333 & 3 \\
\bottomrule
\end{tabular}
}
\caption{Quartile summary for normalized empirical best-layer positions by model family.}
\label{tab:app-family-quartiles}
\end{table}

\begin{table*}[t]
\centering
\scriptsize
\resizebox{\textwidth}{!}{\begin{tabular}{llllrrrr}
\toprule
Family & Training label & Task & Readout & N & Median peak & Early peak & Early 1pp \\
\midrule
Encoder-decoder & Span Corruption & Classification & last token & 18 & 0.08 & 100.0 & 100.0 \\
Encoder-decoder & Span Corruption & Classification & mean pooled tokens & 18 & 0.08 & 100.0 & 100.0 \\
Encoder-decoder & Span Corruption & STS & last token & 6 & 0.08 & 100.0 & 100.0 \\
Encoder-decoder & Contrastive Sentence Embedding Finetuning & Classification & mean pooled tokens & 12 & 0.08 & 91.7 & 100.0 \\
Decoder & Causal Language Modeling & Multiple-choice QA & CLS token & 24 & 0.42 & 45.8 & 100.0 \\
Encoder-decoder & Contrastive Sentence Embedding Finetuning & Multiple-choice QA & last token & 12 & 0.79 & 33.3 & 100.0 \\
Encoder-decoder & Denoising Autoencoding & Multiple-choice QA & last token & 6 & 0.83 & 16.7 & 100.0 \\
Encoder & Knowledge Distillation + MLM & Classification & mean pooled tokens & 8 & 0.17 & 62.5 & 87.5 \\
Encoder & Continual Multi-Task Pretraining & Classification & mean pooled tokens & 8 & 0.17 & 62.5 & 87.5 \\
Encoder & Masked Language Modeling (MLM) & Classification & mean pooled tokens & 56 & 0.32 & 53.6 & 87.5 \\
Encoder-decoder & Span Corruption & Multiple-choice QA & last token & 18 & 0.38 & 50.0 & 77.8 \\
Encoder & Whole Word Masking (WWM) & Classification & mean pooled tokens & 8 & 0.08 & 62.5 & 75.0 \\
Encoder & Span Masking + Span Boundary Objective & Classification & mean pooled tokens & 8 & 0.17 & 62.5 & 75.0 \\
Encoder & Replaced Token Detection (RTD) & Classification & mean pooled tokens & 8 & 0.33 & 50.0 & 75.0 \\
Encoder & Span Masking + Span Boundary Objective & Pair classification & sentence-pair readout & 6 & 0.38 & 50.0 & 66.7 \\
Encoder-decoder & Contrastive Sentence Embedding Finetuning & Classification & last token & 12 & 0.83 & 33.3 & 66.7 \\
\bottomrule
\end{tabular}
}
\caption{Representative peak-depth patterns by model family, training label, task, and readout.  ``Median peak'' is median normalized best-layer depth.  ``Early peak'' is the percentage of exact best layers in the first third.  ``Early 1pp'' is the percentage of groups whose earliest layer within 0.01 of best is in the first third.}
\label{tab:app-main-strata}
\end{table*}

\begin{table*}[t]
\centering
\scriptsize
\resizebox{\textwidth}{!}{\begin{tabular}{llllrrr}
\toprule
Family & Training label & Task & Readout & N & Early 1pp & Gap \\
\midrule
Encoder-decoder & Span Corruption & Classification & last token & 18 & 100.0 & 0.185 \\
Encoder-decoder & Span Corruption & Classification & mean pooled tokens & 18 & 100.0 & 0.295 \\
Encoder-decoder & Span Corruption & STS & last token & 6 & 100.0 & 0.191 \\
Encoder-decoder & Contrastive Sentence Embedding Finetuning & Classification & mean pooled tokens & 12 & 100.0 & 0.093 \\
Decoder & Causal Language Modeling & Multiple-choice QA & CLS token & 24 & 100.0 & 0.005 \\
Encoder-decoder & Contrastive Sentence Embedding Finetuning & Multiple-choice QA & last token & 12 & 100.0 & 0.003 \\
Encoder-decoder & Denoising Autoencoding & Multiple-choice QA & last token & 6 & 100.0 & 0.005 \\
Encoder & Knowledge Distillation + MLM & Classification & mean pooled tokens & 8 & 87.5 & 0.017 \\
Encoder & Continual Multi-Task Pretraining & Classification & mean pooled tokens & 8 & 87.5 & 0.026 \\
Encoder & Masked Language Modeling (MLM) & Classification & mean pooled tokens & 56 & 87.5 & 0.036 \\
Encoder-decoder & Span Corruption & Multiple-choice QA & last token & 18 & 77.8 & 0.005 \\
Encoder & Whole Word Masking (WWM) & Classification & mean pooled tokens & 8 & 75.0 & 0.031 \\
Encoder & Span Masking + Span Boundary Objective & Classification & mean pooled tokens & 8 & 75.0 & 0.026 \\
Encoder & Replaced Token Detection (RTD) & Classification & mean pooled tokens & 8 & 75.0 & 0.081 \\
\bottomrule
\end{tabular}
}
\caption{Highest early near-best plateau groups among groups with at least six instances.}
\label{tab:app-top-early}
\end{table*}

\begin{table*}[t]
\centering
\scriptsize
\resizebox{\textwidth}{!}{\begin{tabular}{llrr}
\toprule
Task & Training label & N & Early near-best (\%) \\
\midrule
Classification & Span Corruption & 36 & 100.0 \\
Classification & Masked Language Modeling (MLM) & 111 & 60.4 \\
Classification & Contrastive Sentence Embedding Finetuning & 48 & 56.2 \\
Classification & Continual Multi-Task Pretraining & 16 & 56.2 \\
Classification & Knowledge Distillation + MLM & 16 & 43.8 \\
Classification & Replaced Token Detection (RTD) & 16 & 43.8 \\
Classification & Whole Word Masking (WWM) & 16 & 37.5 \\
Classification & Span Masking + Span Boundary Objective & 16 & 37.5 \\
Classification & Denoising Autoencoding & 12 & 25.0 \\
Multiple-choice QA & Contrastive Sentence Embedding Finetuning & 12 & 100.0 \\
Multiple-choice QA & Denoising Autoencoding & 6 & 100.0 \\
Multiple-choice QA & Span Corruption & 18 & 77.8 \\
Multiple-choice QA & Causal Language Modeling & 48 & 77.1 \\
Multiple-choice QA & Causal Language Modeling (State-Space) & 6 & 33.3 \\
Pair classification & Masked Language Modeling (MLM) & 88 & 42.0 \\
Pair classification & Contrastive Sentence Embedding Finetuning & 6 & 33.3 \\
Pair classification & Span Masking + Span Boundary Objective & 12 & 33.3 \\
Pair classification & Replaced Token Detection (RTD) & 12 & 25.0 \\
Pair classification & Whole Word Masking (WWM) & 16 & 18.8 \\
Pair classification & Knowledge Distillation + MLM & 12 & 16.7 \\
Pair classification & Continual Multi-Task Pretraining & 12 & 16.7 \\
STS & Span Corruption & 6 & 100.0 \\
STS & Contrastive Sentence Embedding Finetuning & 16 & 37.5 \\
STS & Whole Word Masking (WWM) & 8 & 12.5 \\
STS & Masked Language Modeling (MLM) & 32 & 3.1 \\
STS & Denoising Autoencoding & 2 & 0.0 \\
STS & Knowledge Distillation + MLM & 4 & 0.0 \\
STS & Continual Multi-Task Pretraining & 4 & 0.0 \\
\bottomrule
\end{tabular}
}
\caption{Percentages of peak-layer instances whose earliest near-best layer lies in the first third of the model.}
\label{tab:app-early-heatmap-data}
\end{table*}

\subsection{Quadruplet Audit}

The model-family, training-label, task, and readout audit contains 58 observed combinations. Table~\ref{tab:app-quadruplet-count} reports how many remain as the minimum number of supporting instances is increased. 

\begin{table}[h]
\centering
\small
\begin{tabular}{lr}
\toprule
Pruning rule & Count \\
\midrule
All observed quadruplets & 58 \\
Quadruplets with at least 6 instances & 43 \\
Quadruplets with at least 8 instances & 28 \\
Quadruplets with at least 10 instances & 16 \\
\bottomrule
\end{tabular}

\caption{Number of distinct model-family/training-label/task/readout combinations retained at increasing minimum-support thresholds.}
\label{tab:app-quadruplet-count}
\end{table}

\subsection{Matched Pooling Analysis}

\begin{figure}[h]
\centering
\includegraphics[width=\columnwidth]{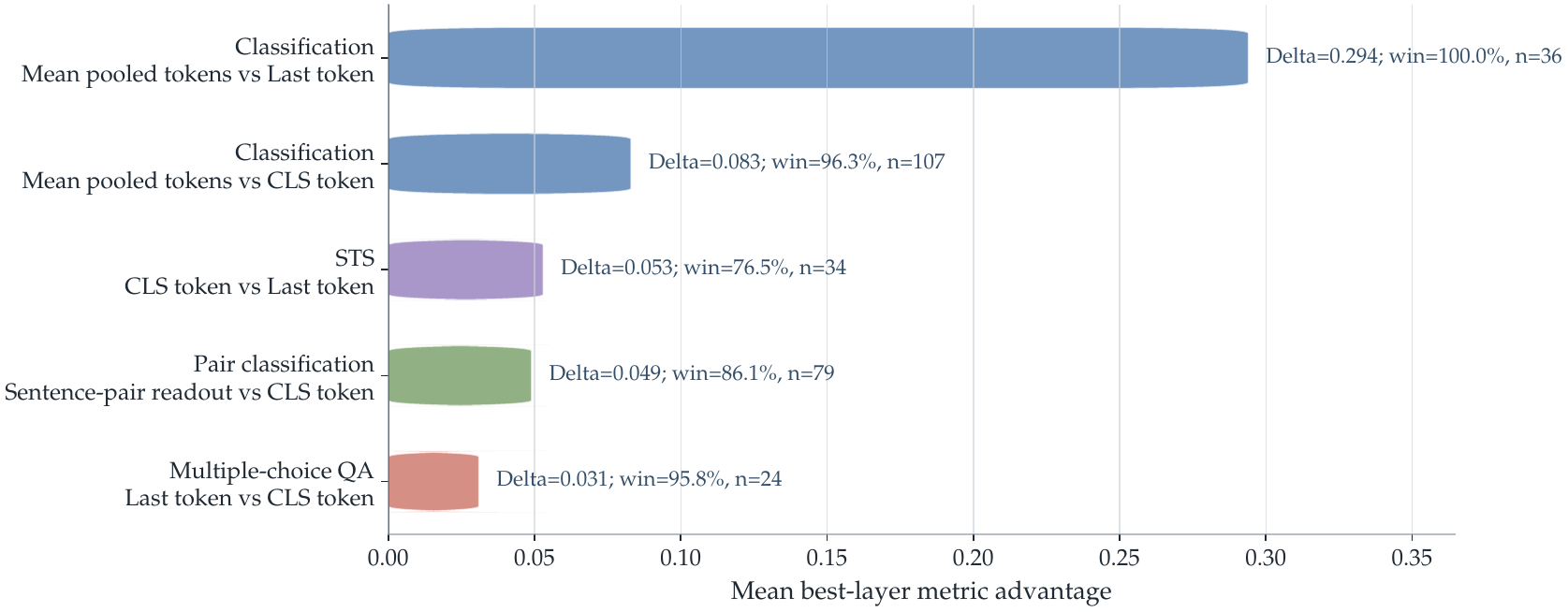}
\caption{Pairwise readout advantages within matched comparison groups.}
\label{fig:app-pooling-pairwise}
\end{figure}

The peak-depth analysis asks where a given readout peaks.  The matched pooling analysis asks which readout attains the best score within groups that share the same model, dataset, task, and probe.  This separates early accessibility from best attainable performance. Detailed matched pooling results are reported in Table~\ref{tab:app-pooling-matched}. Pairwise pooling comparisons are summarized in Table~\ref{tab:app-pooling-pairwise}. Figure~\ref{fig:app-pooling-pairwise} further illustrates pairwise pooling performance differences.

\begin{table*}[t]
\centering
\scriptsize
\resizebox{\textwidth}{!}{\begin{tabular}{llrrrr}
\toprule
Task & Readout & N & Win (\%) & Mean gap & Early 1pp (\%) \\
\midrule
Classification & Mean pooled tokens & 143 & 97.2 & 0.000 & 83.2 \\
Classification & CLS token & 107 & 3.7 & -0.083 & 20.6 \\
Classification & Last token & 36 & 0.0 & -0.294 & 72.2 \\
Multiple-choice QA & Last token & 24 & 95.8 & 0.000 & 54.2 \\
Multiple-choice QA & CLS token & 24 & 4.2 & -0.031 & 100.0 \\
Pair classification & Sentence-pair readout & 79 & 86.1 & -0.002 & 51.9 \\
Pair classification & CLS token & 79 & 13.9 & -0.052 & 15.2 \\
STS & CLS token & 34 & 76.5 & -0.021 & 5.9 \\
STS & Last token & 34 & 23.5 & -0.074 & 8.8 \\
\bottomrule
\end{tabular}
}
\caption{Matched readout comparison by task.  Win rate is computed only within groups that share the same model, dataset, task, and probe type.  Mean gap is the readout's best-layer metric minus the best readout metric in the same matched group.}
\label{tab:app-pooling-matched}
\end{table*}

\begin{table*}[t]
\centering
\scriptsize
\resizebox{\textwidth}{!}{\begin{tabular}{llrrrr}
\toprule
Task & Higher-performing readout vs alternative & N & Win (\%) & Mean delta & Median delta \\
\midrule
Classification & Mean pooled tokens vs CLS token & 107 & 96.3 & 0.083 & 0.060 \\
Classification & Mean pooled tokens vs Last token & 36 & 100.0 & 0.294 & 0.194 \\
Multiple-choice QA & Last token vs CLS token & 24 & 95.8 & 0.031 & 0.013 \\
Pair classification & Sentence-pair readout vs CLS token & 79 & 86.1 & 0.049 & 0.042 \\
STS & CLS token vs Last token & 34 & 76.5 & 0.053 & 0.014 \\
\bottomrule
\end{tabular}
}
\caption{Pairwise readout comparisons oriented toward the higher-performing readout in each task-specific pair.}
\label{tab:app-pooling-pairwise}
\end{table*}

\subsection{SGI Selector Diagnostics}

\begin{figure}[t]
\centering
\includegraphics[width=\columnwidth]{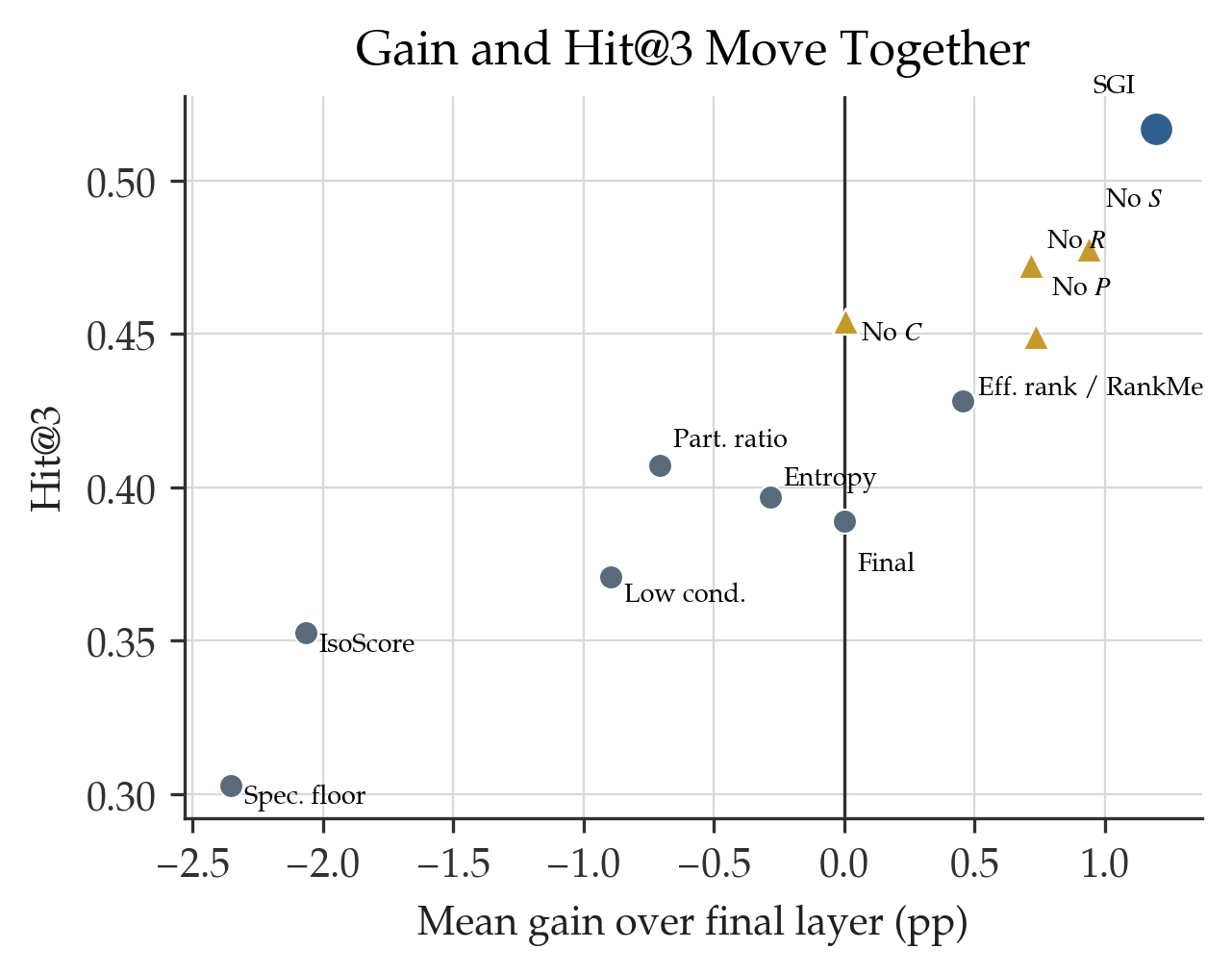}
\caption{Aggregate selector performance on 383 encoder evaluation groups.
Each point is one selector or ablation, plotted by mean gain over the final
layer (x-axis) and Hit@3 against the supervised oracle ranking (y-axis).}
\label{fig:app-sgi-hit3-gain}
\end{figure}

Figure~\ref{fig:app-sgi-hit3-gain} summarizes the encoder selector comparison
from Section~\ref{subsec:ablations} and Table~\ref{tab:ablation} along two
aggregate axes: mean gain over the final layer and Hit@3. The ablation points
correspond to removing one term from
$G_\ell = R_\ell + C_\ell + S_\ell + P_\ell$: No $R$ uses $C+S+P$, No $C$ uses
$R+S+P$, No $S$ uses $R+C+P$, and No $P$ uses $R+C+S$. Thus the figure shows
both how SGI compares to single-signal label-free baselines and how each SGI
coordinate contributes to the full selector. SGI lies in the upper-right region:
it has the largest mean gain over the final layer and the highest Hit@3 among
the plotted label-free methods and ablations. This supports the main result
that the full index improves not only average performance but also the chance
of selecting a layer close to the supervised oracle.

\begin{figure}[t]
\centering
\includegraphics[width=\columnwidth]{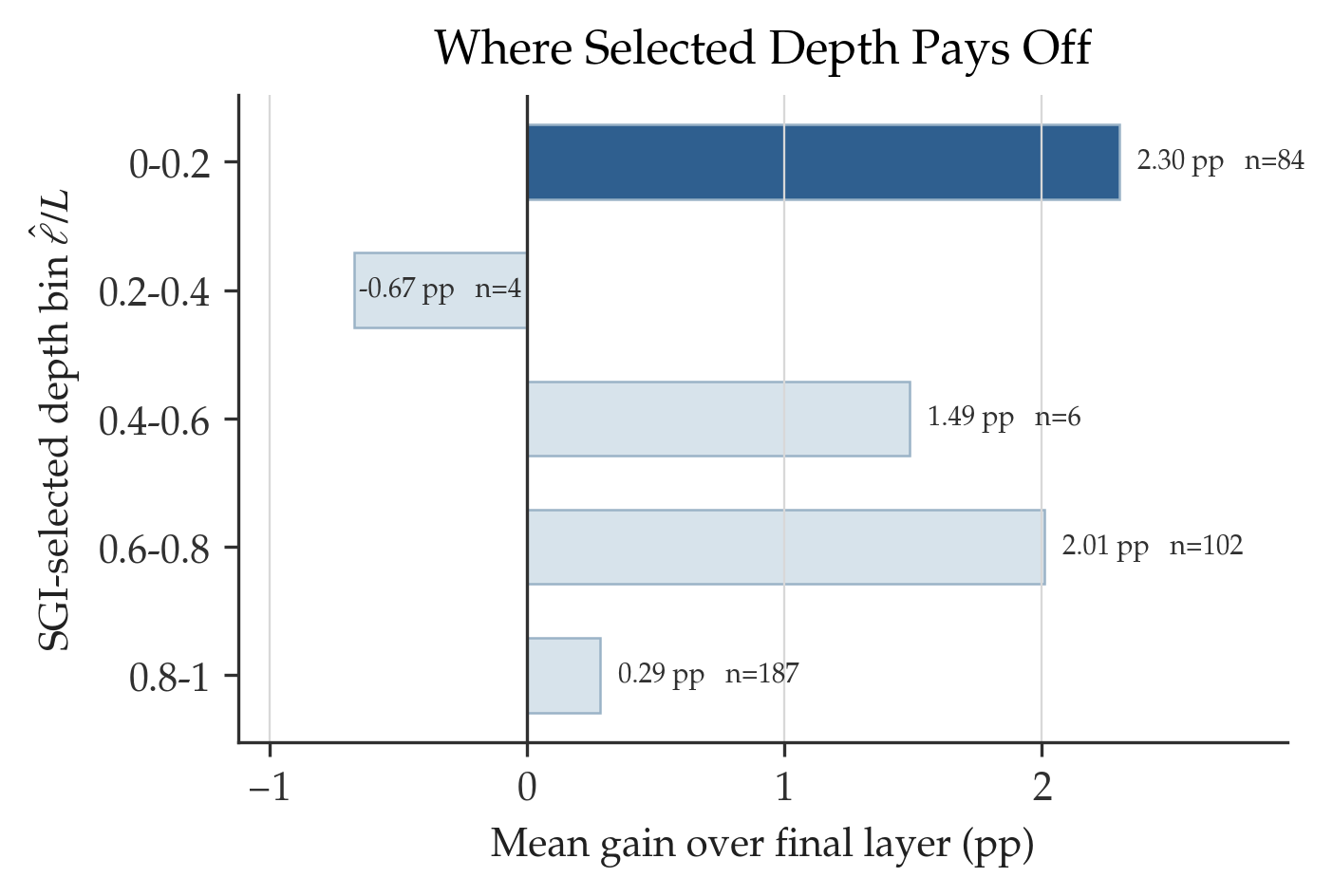}
\caption{SGI gain by selected-layer depth. Groups are binned by normalized
selected depth $\hat{\ell}/L$; bars show mean gain over the final layer, with
group counts printed next to each bar.}
\label{fig:app-sgi-depth-bin}
\end{figure}

Figure~\ref{fig:app-sgi-depth-bin} stratifies the full SGI result by the
normalized depth of the selected layer. The plot is descriptive: it asks where
SGI tends to obtain gains relative to the final layer. Near-final selections
are most frequent, but their average gain is modest. The larger gains appear
when SGI selects very early layers or late-middle layers, indicating that the
method is not simply reproducing the final-layer default. The small negative
bar in the $0.2$--$0.4$ bin should be interpreted cautiously because it contains
only four evaluation groups.

\subsection{Geometry Evidence And Overlays}

This subsection documents the raw geometric evidence underlying SGI and connects each coordinate to its representational interpretation.  Recall that SGI treats each layer $\ell$ as a point cloud $H_\ell \in \mathbb{R}^{n_\ell \times d}$ and summarises its shape through four coordinates: spectral capacity $e_\ell$, power-law fit quality $x_\ell$, condition number $\kappa_\ell$, and local intrinsic dimensionality $\delta_\ell$.  The four coordinates probe complementary geometric properties of the representation manifold, and their roles are as follows.

\paragraph{Spectral capacity and effective rank.}
The effective rank $e_\ell = \exp(-\sum_i \pi_{\ell,i} \log \pi_{\ell,i})$, where $\pi_{\ell,i} = \sigma_{\ell,i}/\sum_j \sigma_{\ell,j}$, quantifies how uniformly the singular-value mass is distributed across directions.  A layer with high effective rank occupies many orthogonal directions in the ambient space, indicating that its representations are \emph{isotropically spread} in the sense of not collapsing onto a low-dimensional subspace.  By contrast, layers with collapsed spectra --- characteristic of final layers in contextual encoders \citep{ethayarajh2019contextual} --- concentrate variance along a few dominant axes, leaving the bulk of the ambient space unused.  Table~\ref{tab:app-geometry-features} reports the log-min-max normalised effective rank $\Lup(e)_\ell$ alongside the raw SGI score; Table~\ref{tab:app-geometry-delta} reports the signed difference between the best-layer coordinate and the final-layer coordinate, making it possible to verify directly which geometric dimensions favour intermediate layers over the endpoint.

\paragraph{Power-law fit quality and spectral residual structure.}
A pure power-law spectrum $\sigma_{\ell,i} \propto i^{-\alpha}$ is the signature of maximum-entropy heavy-tailed self-regularisation predicted by random-matrix theory \citep{martin2021implicitselfreg}.  The coefficient of determination $x_\ell$ measures how well this single-parameter law accounts for the observed singular values: high $x_\ell$ implies that the spectrum is smooth and featureless, while low $x_\ell$ indicates structured deviation --- residual bumps, kinks, or plateaus --- that are not captured by the power law alone.  Crucially, our ablations (Section~\ref{subsec:ablations}) show that $x_\ell$ carries more layer-quality signal than the exponent $\alpha$ itself, reversing a common assumption in the spectral-decay literature.  The residual term $R_\ell = \Ndown(x)_\ell^2$ therefore rewards layers whose spectra \emph{depart} from the idealised heavy-tailed shape, a regime that empirically coincides with the transition between lexical and contextual representation phases in intermediate depth.

\paragraph{Condition number and isometric stability.}
The condition number $\kappa_\ell = \sigma_{\ell,1}/\sigma_{\ell,q}$ is the ratio of the largest to the smallest non-zero singular value. Weyl's inequality bounds singular-value changes under an additive perturbation, $|\sigma_i(H_\ell+\Delta H)-\sigma_i(H_\ell)|\leq\|\Delta H\|_2$; it does not by itself bound fitted probe weights or guarantee generalisation. A small non-zero singular value can make an unregularised inverse problem sensitive to noise, while regularisation, the loss, and labels also affect probe stability. Large condition numbers can indicate the rogue-dimension and anisotropy phenomena discussed by \citet{timkey-van-schijndel-2021-bark}. SGI uses conditioning as a complementary diagnostic of spectral degeneracy, rather than a sufficient condition for downstream separability or preservation of input-space distances.

\paragraph{Local intrinsic dimensionality and manifold spread.}
The TwoNN estimator $\delta_\ell$ \citep{Facco2017} recovers the intrinsic dimensionality of the data manifold from the ratio of first- to second-nearest-neighbour distances under a locally Poisson assumption.  Unlike global spectral capacity, which describes how variance is allocated across orthogonal axes, $\delta_\ell$ describes how points populate the manifold in their immediate neighbourhood, providing a complementary local view of manifold geometry.  A layer with high $\delta_\ell$ places its representations on a locally high-dimensional manifold, which implies that the neighbourhood structure of the input space is preserved in many directions simultaneously --- a geometric isometry property at the local scale.  Layers where $\delta_\ell$ is low tend to project inputs onto a low-dimensional attractor, compressing fine-grained distinctions that a linear probe would otherwise exploit.

\begin{figure*}[t]
\centering
\includegraphics[width=.95\textwidth]{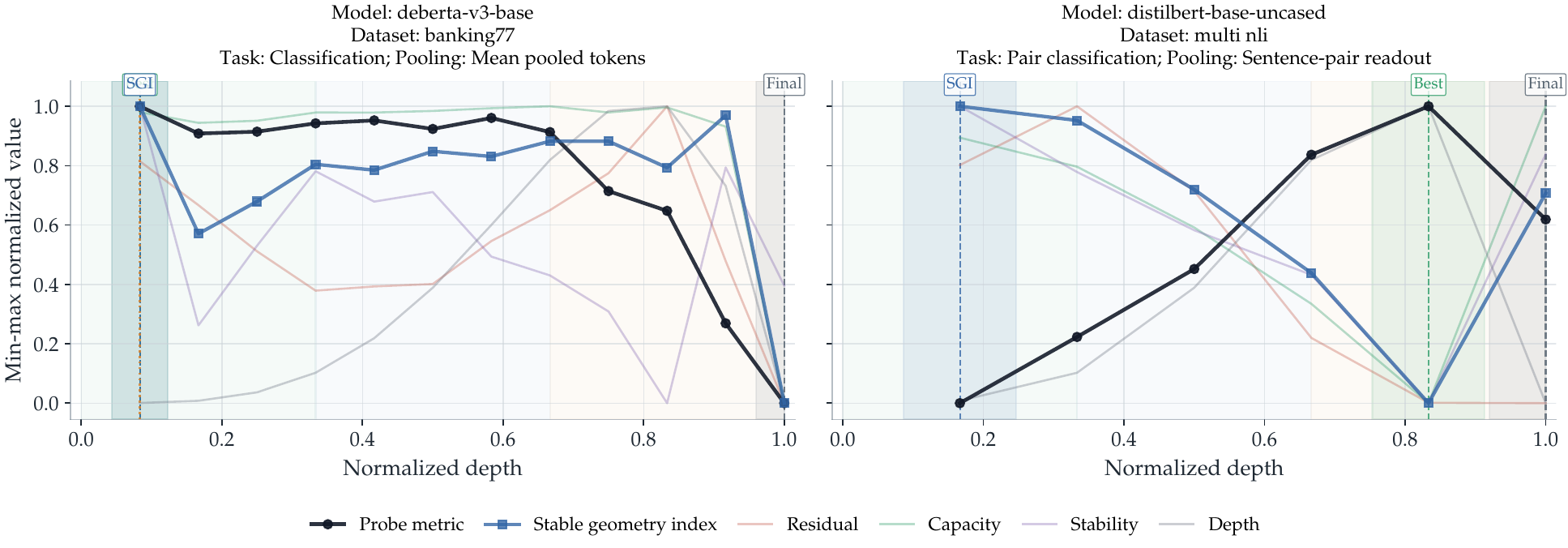}
\caption{Examples for early-layer behavior and selector caution.  The left panel is an early near-best case; the right panel is a boundary case where the geometry curve should not be read as a universal proxy for probe performance.}
\label{fig:app-overlay-caution}
\end{figure*}

\begin{figure*}[t]
\centering
\includegraphics[width=.95\textwidth]{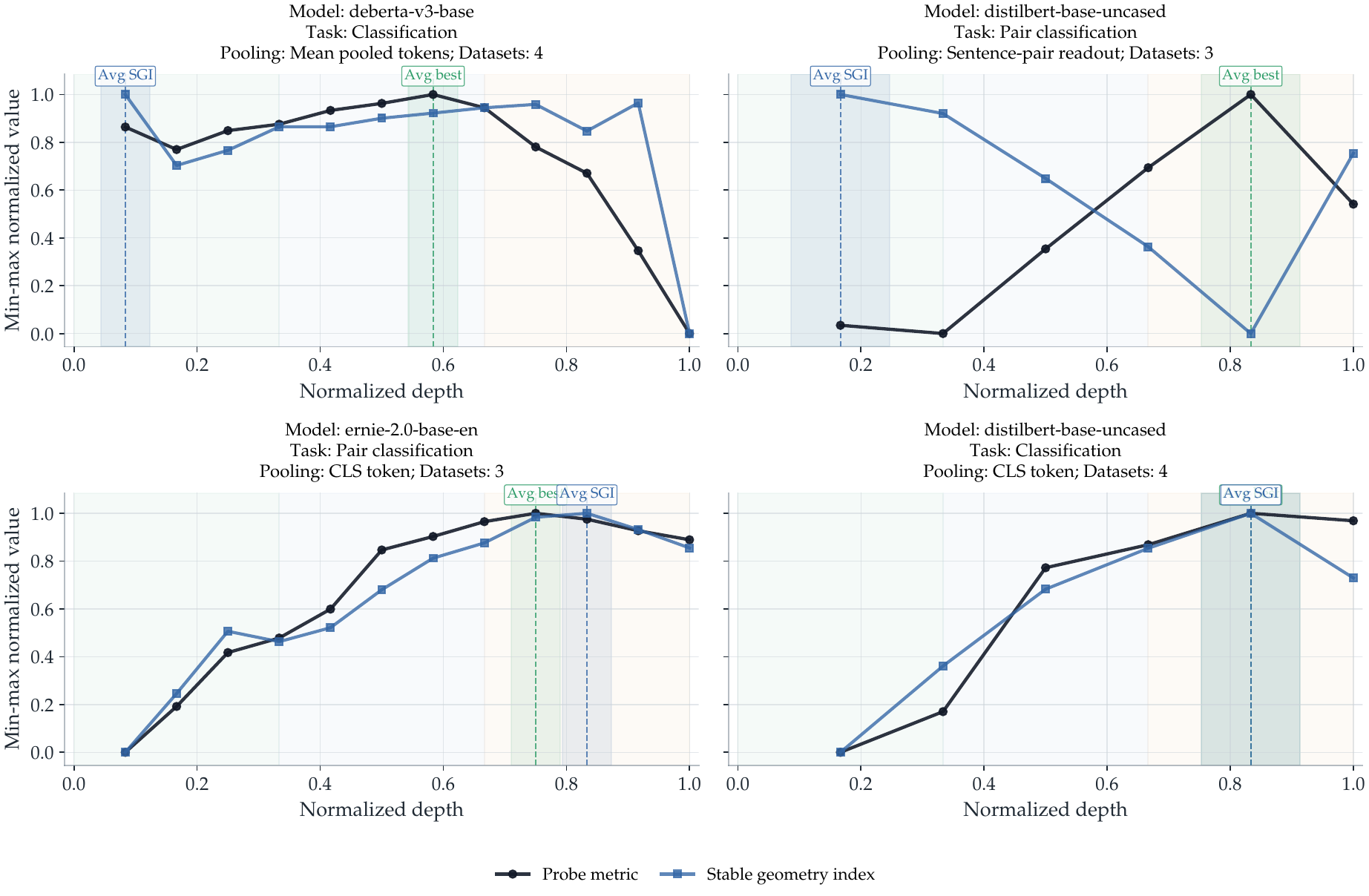}
\caption{Overlay candidates averaged across datasets.  The main-candidate panels provide model-level qualitative support, while the appendix and caution panels show boundary behavior.}
\label{fig:app-model-avg-overlays}
\end{figure*}

\paragraph{Reading the tables and overlays.}
Table~\ref{tab:app-geometry-features} lists the final normalised SGI coordinates for each evaluation group; these are the quantities that SGI aggregates additively to produce the selected layer.  Table~\ref{tab:app-geometry-delta} gives the signed difference (best layer minus final layer) for each coordinate, allowing a direct audit of which geometric dimensions are higher at the empirically best layer than at the endpoint.  Positive deltas for capacity, residual structure, and intrinsic spread indicate that the best layer is geometrically richer than the final layer along those axes, while a positive delta for stability (lower condition number) indicates that it is also more isometrically well-conditioned.  The overlay figures are qualitative checks against the probe curves: the main panels (Figure~\ref{fig:app-overlay-positive}) show cases where the SGI curve tracks the probe performance curve and the selected layer falls in the top-three probe layers, while the caution panels (Figure~\ref{fig:app-overlay-caution}) show boundary cases where the geometry curve diverges from supervised performance, confirming that SGI should be read as a phase detector rather than a universal proxy for probe performance. Quantitative overlay statistics are provided in Table~\ref{tab:app-model-avg-overlays}. Figure~\ref{fig:app-model-avg-overlays} shows averaged overlay trends across models.

\begin{table*}[t]
\centering
\scriptsize
\resizebox{\textwidth}{!}{\begin{tabular}{llll}
\toprule
Coordinate & Formula & Source & Interpretation \\
\midrule
Residual & $R=\operatorname{lo}(\alpha_{\mathrm{req}}^2)^2$ & $\alpha_{\mathrm{req}}^2$ & Low residual power-law error \\
Capacity & $C=Z(1+Z)$ & effective rank & High spectral capacity \\
Stability & $S=\operatorname{lo}(\kappa)+\operatorname{ranklo}(\kappa)+\operatorname{loghi}(d_{2NN})$ & condition number, 2NN dimension & Stable spread \\
Depth & $P=\operatorname{hi}(u^4(1-u))$ & layer depth & Late-middle phase prior \\
\bottomrule
\end{tabular}
}
\caption{Final geometry coordinates used by SGI.}
\label{tab:app-geometry-features}
\end{table*}

\begin{table*}[t]
\centering
\scriptsize
\resizebox{\textwidth}{!}{\begin{tabular}{lrrrrr}
\toprule
Coordinate & Task & N & Mean delta & Median delta & Std. \\
\midrule
Capacity & Classification & 191 & 0.152 & 0.000 & 0.831 \\
Capacity & Pair classification & 144 & -0.220 & -0.187 & 0.757 \\
Capacity & STS & 48 & -0.717 & -0.497 & 0.769 \\
Depth & Classification & 191 & 0.255 & 0.003 & 0.369 \\
Depth & Pair classification & 144 & 0.540 & 0.606 & 0.412 \\
Depth & STS & 48 & 0.709 & 0.819 & 0.309 \\
Residual & Classification & 191 & 0.081 & 0.000 & 0.468 \\
Residual & Pair classification & 144 & 0.001 & 0.000 & 0.471 \\
Residual & STS & 48 & -0.230 & -0.233 & 0.395 \\
Stability & Classification & 191 & 0.210 & 0.000 & 0.984 \\
Stability & Pair classification & 144 & -0.089 & 0.000 & 0.817 \\
Stability & STS & 48 & -0.169 & -0.081 & 0.632 \\
Stable geometry index & Classification & 191 & 0.699 & 0.169 & 1.816 \\
Stable geometry index & Pair classification & 144 & 0.231 & 0.058 & 1.446 \\
Stable geometry index & STS & 48 & -0.407 & -0.278 & 1.181 \\
\bottomrule
\end{tabular}
}
\caption{Per-coordinate difference between the empirical best layer and the final layer. Positive values indicate the best layer is richer than the endpoint along that geometric axis.}
\label{tab:app-geometry-delta}
\end{table*}

\begin{table*}[t]
\centering
\scriptsize
\resizebox{\textwidth}{!}{\begin{tabular}{lllrrrrl}
\toprule
Model & Task & Readout & Datasets & Best & Selected & Spearman & Verdict \\
\midrule
deberta-v3-base & Classification & Mean pooled tokens & 4 & 7 & 1 & 0.315 & appendix candidate \\
distilbert-base-uncased & Pair classification & Sentence-pair readout & 3 & 5 & 1 & -0.886 & caution only \\
ernie-2.0-base-en & Pair classification & CLS token & 3 & 9 & 10 & 0.972 & main candidate \\
distilbert-base-uncased & Classification & CLS token & 4 & 5 & 5 & 0.943 & main candidate \\
\bottomrule
\end{tabular}
}
\caption{Model-level overlay analysis after averaging across datasets.  ``Selected'' is the SGI layer on the averaged curve.}
\label{tab:app-model-avg-overlays}
\end{table*}

\section{Artifact, Data, and Reproducibility Statement}
\label{app:artifact-reproducibility}

This appendix documents the scientific artifacts used and created in this work, their intended use, data and model coverage, experimental settings, and reporting conventions. The study uses public pretrained checkpoints and public benchmark datasets accessed through their upstream distribution channels. The artifacts created by this work are the Stable Geometry Index (SGI) selection rule, configuration files, layer-selection outputs, aggregate tables, and figures. We do not redistribute upstream model weights or raw dataset text; users should obtain those artifacts from their original providers and follow the corresponding license and access terms.

\subsection{Scientific Artifacts and Licenses}
\label{app:artifact-licenses}

We use publicly released pretrained checkpoints and public benchmark datasets obtained from their original sources. We do not redistribute upstream model weights or raw dataset text; users should follow the original license and access terms. Where license metadata was missing or unclear in the queried cards, we treated the upstream source as authoritative and used the artifact only for research evaluation. A summary of dataset and artifact licenses is provided in Table~\ref{tab:artifact-license-summary}.

\begin{table*}[t]
\centering
\small
\setlength{\tabcolsep}{4pt}
\begin{tabular}{p{0.13\textwidth}p{0.23\textwidth}p{0.55\textwidth}}
\toprule
Artifact class & License or access condition observed & Artifacts \\
\midrule
Models & Apache-2.0 & BERT variants, DistilBERT, ModernBERT, MPNet sentence encoder, Pythia-70M, Qwen2.5-0.5B, T5, BART, Flan-T5, ByT5, Instructor, Sentence-T5 \\
Models & MIT & RoBERTa, DeBERTa-v3, XLM-RoBERTa, BGE, GPT-2, GPT-Neo \\
Models & CC-BY-NC-4.0 & XLM-MLM-100-1280 \\
Models & Not specified in queried metadata & SpanBERT, ERNIE-2.0-base-en, Mamba-130M \\
\midrule
Datasets & MIT & Banking77 \\
Datasets & CC-BY-SA-4.0 & SNLI, ARC-Challenge \\
Datasets & Mixed metadata & MultiNLI \\
Datasets & Other or unknown metadata & AG News, Emotion, SST-2, MRPC, QQP, STS-B, BIOSSES, HellaSwag, PIQA \\
\bottomrule
\end{tabular}
\caption{License and access-condition summary for the upstream artifacts. Missing license metadata means that the queried artifact card did not expose a single clear license field; it does not imply that the artifact is unrestricted.}
\label{tab:artifact-license-summary}
\end{table*}

\subsection{Intended Use}
\label{app:intended-use}

All upstream artifacts are used for offline research benchmarking, representation analysis, and layer-selection evaluation. This use is consistent with the common intended use of pretrained checkpoints as research representation models and of benchmark datasets as evaluation resources. SGI and the derived tables are intended for research on label-free layer selection and representation geometry. They are not intended for deployment in user-facing systems, profiling individuals, making decisions about people, or replacing task-specific validation in safety-critical applications. Any derived embeddings or geometry files should be treated as research artifacts that inherit the original access conditions of the corresponding upstream datasets and models.

\subsection{Privacy, Identifying Information, and Offensive Content}
\label{app:privacy-content}

We did not collect new human-subject data. The study uses existing public benchmarks whose examples may contain naturally occurring names, web text, social text, biomedical text, news text, or potentially offensive language inherited from their original sources. We checked dataset cards, schemas, task descriptions, and available metadata to identify fields containing free-form text and labels. The experiments do not require identity inference, demographic annotation, or publication of raw examples. The paper reports only aggregate metrics, layer indices, geometry summaries, and figures. If derived embeddings or per-example outputs are released, they should be distributed only under research-use conditions and should not be treated as anonymized raw data, because dense representations may still retain information about their source examples.

\subsection{Artifact Documentation and Coverage}
\label{app:artifact-coverage}

The broad layerwise study covers 25 pretrained checkpoints, 13 datasets, four task categories, four readout strategies, and two probe families. Sentence-transformer checkpoints are treated as encoder-family models because they expose encoder-style sentence representations. The evaluated task datasets are English. The model set includes monolingual English models, multilingual encoders, decoder-only language models, encoder--decoder models, sentence-transformer models, and one state-space model. The task coverage includes sentiment, news, emotion, banking intent, natural-language inference, paraphrase detection, question-pair classification, semantic textual similarity, science question answering, commonsense sentence completion, and physical commonsense reasoning. Demographic groups are not explicitly annotated in the reported benchmark inventory, and the paper does not make demographic performance claims. Model family parameter counts are listed in Table~\ref{tab:model-family-params}.

\begin{table*}[t]
\centering
\small
\setlength{\tabcolsep}{4pt}
\begin{tabular}{p{0.17\textwidth}p{0.34\textwidth}rrrr}
\toprule
Family & Examples & Count & Min params & Max params & Layer range \\
\midrule
Encoder & BERT, RoBERTa, DeBERTa, XLM-R & 12 & 66.4M & 571.5M & 6--24 \\
Sentence transformer & MPNet, BGE & 2 & 109.5M & 109.5M & 12 \\
Decoder & GPT-2, GPT-Neo, Pythia, Qwen2.5 & 4 & 70.4M & 494.0M & 6--24 \\
Encoder--decoder & T5, BART, Flan-T5, ByT5, Instructor, Sentence-T5 & 6 & 139.4M & 581.1M & 6--18 \\
State-space & Mamba-130M & 1 & 129.1M & 129.1M & 24 \\
\bottomrule
\end{tabular}
\caption{Model-family coverage and parameter scale. Parameter counts use unique tied parameters after applying model weight tying. The full range is 66.4M parameters for DistilBERT-base to 581.1M parameters for ByT5-base.}
\label{tab:model-family-params}
\end{table*}

\subsection{Dataset Statistics}
\label{app:data-statistics}

Table~\ref{tab:dataset-statistics} reports the dataset sizes observed from the Hugging Face dataset builders and the configured splits used by the experiments. When a labeled official test split was configured, probes train on the training split and evaluate on the official test split. When no suitable labeled test split was configured, the pipeline uses a deterministic 60/20/20 train/validation/test split of the labeled split with seeds 42, 1337, and 2024. The broad study contains 617 model--dataset--task--readout--probe groups; the main SGI encoder selection result contains 383 encoder groups.

\begin{table*}[t]
\centering
\footnotesize
\setlength{\tabcolsep}{3pt}
\begin{tabular}{llrrrr}
\toprule
Dataset & Task & Train & Dev & Test & Groups \\
\midrule
AG News & Classification & 120000 & -- & 7600 & 80 \\
Emotion & Classification & 16000 & 2000 & 2000 & 79 \\
Banking77 & Classification & 9993 & -- & 3076 & 80 \\
SST-2 & Classification & 67349 & 872 & 1821 & 48 \\
MRPC & Pair classification & 3668 & 408 & 1725 & 52 \\
QQP & Pair classification & 363846 & 40430 & 390965 & 10 \\
MultiNLI & Pair classification & 392702 & 9815 & -- & 48 \\
SNLI & Pair classification & 550152 & 10000 & 10000 & 48 \\
STS-B & STS & 5749 & 1500 & 1379 & 70 \\
BIOSSES & STS & -- & -- & 100 & 12 \\
ARC-Challenge & Multiple-choice QA & 1119 & 299 & 1172 & 30 \\
HellaSwag & Multiple-choice QA & 39905 & 10042 & 10003 & 30 \\
PIQA & Multiple-choice QA & 16113 & 1838 & 3084 & 30 \\
\bottomrule
\end{tabular}
\caption{Dataset statistics for the 13 task datasets in the broad study. Groups counts are the evaluated SGI-compatible model--dataset--task--readout--probe settings.}
\label{tab:dataset-statistics}
\end{table*}

\subsection{Model Size, Compute Budget, and Infrastructure}
\label{app:compute}

The main-cohort checkpoints range from 66.4M to 581.1M parameters; the additional scaling probe extends to 4B parameters. Experiments were run on a single NVIDIA RTX A4500 GPU (20 GB) with CUDA 13.0, PyTorch 2.11.0, and mixed-precision embedding extraction. The full study took approximately 66.5 GPU-hours, comprising about 50.1 hours of probe training, 14.2 hours of embedding extraction, and 2.2 hours of geometry computation, with the remainder spent on layer selection and plotting.

\subsection{Experimental Setup and Hyperparameters}
\label{app:hyperparameters}

Representations are extracted from all non-embedding layers for each model, with maximum sequence length 512, truncation enabled, longest-sequence padding within batch, extraction batch size 96, and float16 mixed precision when running on GPU. Four readouts are used when compatible with the model and task: CLS, last token, mean pooling over non-padding tokens, and sentence-pair pooling. Geometry metrics are computed on pooled representations using a 10\% sample of each split, random seed 42, and 100000 random pairs for pairwise anisotropy estimates. Probe evaluation uses linear probes and two-layer nonlinear probes. Nonlinear probes use a hidden width equal to 0.5 times the input dimension and a GELU activation. Probes are trained with AdamW, learning rate \(10^{-3}\), weight decay 0.01, linear warmup ratio 0.1, batch size 128, evaluation batch size 128, gradient clipping at 1.0, early-stopping patience 3, and a maximum of 20 epochs. No hyperparameter search is used for SGI; the same fixed selector is applied across tasks, readouts, splits, and model families.

\subsection{Descriptive Statistics and Reporting}
Unless stated otherwise, all reported selector scores are means over evaluation groups rather than best-of-run values. Hit@1 and Hit@3 are averaged over groups as Boolean indicators. The main encoder result uses 383 groups; the broader analysis uses 617 groups.
\subsection{Software Packages and Settings}
\label{app:software}

The implementation uses PyTorch 2.11.0, Transformers 5.8.0, Datasets 4.8.5, Hugging Face Hub 1.14.0, Sentence-Transformers 5.4.1, NumPy 2.4.4, pandas 3.0.3, scikit-learn 1.8.0, SciPy 1.17.1, Matplotlib 3.10.9, Seaborn 0.13.2, Hydra 1.3.2, OmegaConf 2.3.0, and Safetensors 0.7.0. Tokenization uses the checkpoint-specific Hugging Face tokenizer with max length 512, truncation enabled, and longest padding within each batch. Normalisation in SGI is performed within each model--dataset--task--readout group, so the score does not use global dataset statistics or probe labels.

\subsection{Use of AI Assistance}
\label{app:ai-assistance}

We used AI coding assistants during implementation, specifically for debugging and for standard utilities such as data loading, preprocessing pipelines, and evaluation scripts. All research contributions, experimental design, and conclusions are the authors' own. All AI-generated code was manually reviewed and verified by the authors, who remain responsible for all content, analysis, and final edits.

\end{document}